\documentclass{article}

\usepackage[final]{neurips_2023}

\usepackage[utf8]{inputenc} 
\usepackage[T1]{fontenc}    
\usepackage{hyperref}       
\usepackage{url}            
\usepackage{booktabs}       
\usepackage{amsfonts}       
\usepackage{nicefrac}       
\usepackage{microtype}      
\usepackage{xcolor}         

\usepackage{wrapfig}
\usepackage{subfigure}
\usepackage{pifont}

\usepackage{paralist,amssymb,bbm,mathtools}
\usepackage{multirow}
\usepackage{algorithm,algorithmicx}
\usepackage{algpseudocode}
\usepackage{bm}
\usepackage{float}

\newtheorem{thm}{Theorem}

\newtheorem{lem}{Lemma}

\newtheorem{assume}{Assumption}

\def \E {\mathrm{E}}
\def \R {\mathbb{R}}
\def \D {\mathbb{D}}
\def \I {\mathbb{I}}
\def \P {\mathbb{P}}

\def \A {\mathcal{A}}

\def \ones {\mathbf{1}}
\def \tc {\Gamma}

\def \Var {\mathrm{Var}}

\def \D{\mathcal{D}}
\def \C{\mathcal{C}}
\def \A{\mathbf{A}}
\def \R{\mathbb{R}}

\def \V{\mathbf{V}}
\def \N{\mathbb{N}}
\def \G{\mathcal{G}}

\def \F{\mathcal{F}}

\def \E {\mathrm{E}}
\def \Pr {\mathrm{Pr}}
\def \x{\bm{x}}
\def \u{\bm{u}}

\def \bt{\bm{\theta}}

\DeclareMathOperator*{\argmin}{argmin}
\DeclareMathOperator*{\argmax}{argmax}

\DeclarePairedDelimiter\norm{\lVert}{\rVert}

\title{Efficient Algorithms for Generalized Linear Bandits with Heavy-tailed Rewards}

\author{
  Bo Xue\textsuperscript{\rm 1,2},~~Yimu Wang\textsuperscript{\rm 3},~~Yuanyu Wan\textsuperscript{\rm 4},~~Jinfeng Yi\textsuperscript{\rm 5},~~Lijun Zhang\textsuperscript{\rm 6,7,}\thanks{Lijun Zhang is the corresponding author.}\\
  \textsuperscript{\rm 1}Department of Computer Science, City University of Hong Kong, Hong Kong, China \\
  \textsuperscript{\rm 2}The City University of Hong Kong Shenzhen Research Institute, Shenzhen, China\\
  \textsuperscript{\rm 3}Cheriton School of Computer Science, University of Waterloo, Waterloo, Canada \\
  \textsuperscript{\rm 4}School of Software Technology, Zhejiang University, Ningbo, China \\
  \textsuperscript{\rm 5}JD AI Research, Beijing, China \\
  \textsuperscript{\rm 6}National Key Laboratory for Novel Software Technology, Nanjing University, Nanjing, China \\
  \textsuperscript{\rm 7}Peng Cheng Laboratory, Shenzhen, China\\
  \texttt{boxue4-c@my.cityu.edu.hk},~~\texttt{yimu.wang@uwaterloo.ca},~~\texttt{wanyy@zju.edu.cn} \\
  \texttt{yijinfeng@jd.com},~~\texttt{zhanglj@lamda.nju.edu.cn}
 }

\begin{document}

\maketitle

\begin{abstract}
This paper investigates the problem of generalized linear bandits with heavy-tailed rewards, whose $(1+\epsilon)$-th moment is bounded for some $\epsilon\in (0,1]$. Although there exist methods for generalized linear bandits, most of them focus on bounded or sub-Gaussian rewards and are not well-suited for many real-world scenarios, such as financial markets and web-advertising. To address this issue, we propose two novel algorithms based on truncation and mean of medians. These algorithms achieve an almost optimal regret bound of $\widetilde{O}(dT^{\frac{1}{1+\epsilon}})$, where $d$ is the dimension of contextual information and $T$ is the time horizon. Our truncation-based algorithm supports online learning, distinguishing it from existing truncation-based approaches. Additionally, our mean-of-medians-based algorithm requires only $O(\log T)$ rewards and one estimator per epoch, making it more practical. Moreover, our algorithms improve the regret bounds by a logarithmic factor compared to existing algorithms when $\epsilon=1$. Numerical experimental results confirm the merits of our algorithms.
\end{abstract}

\section{Introduction}\label{intro}
The multi-armed bandits (MAB) is a powerful framework to model the sequential decision-making process with limited information \citep{Robbins:1952}, which has been found applications in various areas such as medical trails \citep{Villar:2015} and advertisement placement \citep{Bandit:suvery}. In the classical $K$-armed bandit problem, an agent selects one of the $K$ arms and receives a reward drawn  independently and identically distributed from an unknown distribution associated with the chosen arm. The goal of the agent is to maximize the cumulative rewards through the trade-off between exploration and exploitation, i.e., pulling the arms that may potentially give better outcomes while also exploiting the knowledge gained from previous trials to select the optimal arm.

One fundamental limitation of MAB is that it ignores contextual information in some scenarios such as advertisement placement \citep{bandit_algorithm:2020}, where features of users and products can provide valuable guidance for decision making. In these cases, decisions should not only rely on rewards from previous epochs but also the contextual information from both past and current epochs. Stochastic Linear Bandits (SLB) has emerged as the most popular model in the last decade to address this limitation, assuming a linear relationship between the contextual vector and the expected reward \citep{Auer:2002,Linear:Bandit:08,Abbasi:2011,Jiachen:2021,Alieva:2021,Yang:2022,He:2022,Bengs:2022}. However, in many real-world applications, such as social network \citep{Filippi:2010}, the assumption of Poisson or logistic relation between the expected reward and contextual vector has demonstrated better performance, which motivates the study of generalized linear bandits (GLB). In each round, the agent first observes a decision set $\D_t\subset\R^d$ composed of contextual vectors. Then, the agent selects an arm $\x_t\in\D_t$ and receives a reward $y_t$ satisfying the expectation,
\begin{equation}\label{expection}
\E[y_t|\x_t]=\mu(\x_t^\top\bt_*)
\end{equation}
where $\bt_*$ is the inherent vector and $\mu(\cdot)$ is the link function, such as the identity function or the logistic function. The performance of the agent is measured by the regret such that
\begin{equation*}
R(T)= \sum_{t=1}^T \left(\mu(\tilde{\x}_t^\top\bt_*)-\mu(\x_t^\top\bt_*)\right)
\end{equation*}
where $\tilde{\x}_t=\argmax_{\x\in\D_t}\mu(\x^\top\bt_*)$ represents the optimal arm in the set $\D_t$.

Extensive research has been conducted on the GLB, with most assuming sub-Gaussian rewards \citep{Filippi:2010,Li:2012,Li:2017,Jun:2017,Lu:2019-2,Zhengyuan：2019,Yuxuan:2021,Chuanhao:2022}. However, it has been observed that in certain sequential decision-making scenarios, such as financial markets \citep{Cont:2000}, the occurrence of extreme returns is much more frequent than the standard normal distribution. This phenomenon is known as heavy-tailed behavior \citep{Foss:2013}, where the existing algorithms are not suitable. To address this limitation, in this study, we focus on the GLB with heavy-tailed rewards \citep{Bubeck:2013}, i.e., the reward obtained at $t$-th round satifies the condition
\begin{equation*}
\E[|y_t|^{1+\epsilon}]\leq u
\end{equation*}
for some $\epsilon\in(0,1]$ and $u>0$. Different from the traditional sub-Gaussian setting, heavy-tailed rewards do not decay exponentially and the estimation of expected rewards is significantly impacted.

\begin{table*}[t]
\centering
\caption{Summary of the existing work for the linear bandits with heavy-tailed rewards. CC is the abbreviation of computational complexity.}
\begin{tabular}{l l l l l l}
\hline
 & Regret & CC\_Truncation & CC\_MoM & Arms & Model \\
\hline
\citet{Medina:2016} & $\widetilde{O}(dT^{\frac{3}{4}})$ &  $O(d^2T)$ & $O(d^2T/\log T)$ & infinite &SLB\\
\citet{Han:2018} & $\widetilde{O}(d\sqrt{T})$ & $O(d^3T+d^2T^2)$ & $O(d^2 T\cdot\log T)$ & infinite&SLB\\
\citet{xue:2020} & $\widetilde{O}(\sqrt{dT})$ & $O(d^2T^2)$ & $O(d^2 T)$ & finite&SLB\\
This work & $\widetilde{O}(d\sqrt{T})$ & $O(d^2T)$ & $O(d^2T/\log T)$ & infinite&GLB\\
\hline
\end{tabular}
\label{table}
\end{table*}


According to the distinguishing characteristic of heavy-tailed distributions where extreme values occur with high probability, previous studies have developed three main strategies to address the issue in parameter estimation \citep{Audibert:2011,Hsu:2014,Zhang:2018,Chowdhury:2019,Lugosi2021,Han:2021,Huang:2022,Diakonikolas:2022,Gorbunov:2022,Kamath:2022,Shaojie:2022,Gou:2023}. One such strategy is truncation \citet{Audibert:2011}, which mitigates the impact of extreme values by truncating them. A recently proposed strategy is the mean of medians approach \citep{Han:2021}, which involves partitioning the samples drawn from the heavy-tailed distribution into multiple groups, taking the median within each group, and computing the mean of these medians. It intuitively reduces the impact of extreme samples, as extreme samples are distributed to both sides, thus the median value is more robust. The third strategy is median of means \citep{Hsu:2014}, which adjusts the order of calculating mean and taking the median in the mean of medians strategy. 

Most existing algorithms for heavy-tailed bandit problems are derived from aforementioned three strategies, with a primary focus on the SLB model \citep{Medina:2016,Han:2018,xue:2020}. To provide a comprehensive overview and facilitate comparison, we present a summary of our results and previous findings on linear bandits with heavy-tailed rewards in Table \ref{table}. For the sake of clarity, the presented regret bounds in Table \ref{table} are under the assumption that the rewards have finite variance. The computational complexity only takes into account multiplication and division operations. Although \citet{Han:2018} and \citet{xue:2020} achieve nearly optimal regret for infinite-armed and finite-armed SLB, respectively, their algorithms are computationally expensive. The latest work utilizing the mean of medians approach demonstrates efficiency but is limited to symmetric rewards \citep{Han:2021}. Therefore, designing efficient heavy-tailed algorithms for GLB with symmetric and asymmetric rewards is an interesting and non-trivial challenge.


Through the delicate employment of heavy-tailed strategies, our contributions to the generalized linear bandit problem with heavy-tailed rewards can be summeraized as follows:
\begin{itemize}
\item We develop two novel algorithms, CRTM and CRMM, which utilize the truncation strategy and mean of medians strategy, respectively. Both algorithms exhibit a sublinear regret bound of $\widetilde{O}(dT^{\frac{1}{1+\epsilon}})$, which is amolst optimal as the lower bound is $\Omega(dT^{\frac{1}{1+\epsilon}})$ \citep{Han:2018}.

\item CRTM reduces the computational complexity from $O(T^2)$ to $O(T)$ when compared to existing truncation-based algorithms \citep{Han:2018,xue:2020}, while CRMM reduces the number of estimator required per round from $O(\log T)$ to only one, as compared to existing median-of-means-based algorithms \citep{Han:2018,xue:2020}.

\item When $\epsilon = 1$, the regret bounds of CRTM and CRMM improves a logarithmic factor of order $\frac{1}{2\alpha}$ and $\frac{1}{2\alpha}-\frac{1}{2}$ for some $\alpha\in(0,1)$, respectively, over the recently proposed method of \citet{Han:2021}\footnotemark. Notably, CRTM extends the method of \citet{Han:2021} from symmetric rewards to general case, making CRTM more practical.

\item We conduct numerical experiments to demonstrate that our proposed algorithms not only achieve a lower regret bound but also require fewer computational resources when applied to heavy-tailed bandit problems.
\end{itemize}

\footnotetext{For $\epsilon_1>\epsilon_2>0$, if the $(1+\epsilon_1)$-th moment of rewards exists, then the $(1+\epsilon_2)$-th moment of rewards is bounded \citep{xue:2020}. Thus, CRTM and CRMM achieve this regret improvement when $\epsilon\geq1$.}

\section{Related Work}\label{sec:1}
In this section, we briefly review the related work on linear bandits. Through out the paper, the $p$-norm of a vector $\x\in\R^d$ is $\norm{\x}_p=(|x_1|^p+\ldots+|x_d|^p)^{1/p}$. Given a positive definite matrix $\A\in\R^{d\times d}$, the weighted Euclidean norm of the vector $\x$ is $\norm{\x}_\A=\sqrt{\x^\top \A\x}$.

\subsection{Generalized Linear Bandits}\label{related-GLB}

\citet{Filippi:2010} was the first to address the generalized linear bandit problem and proposed an algorithm with a regret bound of $\widetilde{O}(d\sqrt{T})$. However, their algorithm is not efficient as it requires storing all the action-feedback pairs encountered so far and performing maximum likelihood estimation at each step. A notable improvement was presented by \citet{Zhang:2016} with the introduction of an efficient algorithm called OL$^2$M, whose space and time complexity at each epoch does not grow over time and achieves a $\widetilde{O}(d\sqrt{T})$ regret. However, their algorithm is limited to the logistic link function. Later, \citet{Jun:2017} extended OL$^2$M to generic link functions while still maintaining the $\widetilde{O}(d\sqrt{T})$ regret bound. \cite{Ding:2021} proposed another efficient generalized linear bandit algorithm following the line of Thompson sampling scheme.

The main challenge in the bandit problem is the trade-off between exploration and exploitation. To address this issue, the most commonly used approach is the confidence-region-based method, specifically for the linear bandit model with infinite arms \citep{Linear:Bandit:08,Abbasi:2011,Zhang:2016}. Here we take the algorithm OL$^2$M to give a brife introduction to this approach \citep{Zhang:2016}. With the arrival of a new trial $(\x_t,y_t)$ in the $t$-th epoch, OL$^2$M first constructs a surrogate loss $\ell_t(\bt)$ satisfying $\nabla\ell_t(\bt)=(-y_t+\mu(\x_t^\top\bt))\x_t$. Then, OL$^2$M employs a variant of the online Newton step (ONS) to update the estimated parameters, i.e.,
\begin{equation}\label{ONS}
\hat{\bt}_{t+1}^{N}=\argmin_{\bt\in\R^d}\frac{\norm{\bt-\hat{\bt}_t^{N}}^2_{\V_{t+1}}}{2}+\langle\bt-\hat{\bt}_t^{N},\nabla\ell_t(\hat{\bt}_t^{N})\rangle.
\end{equation}
Here, $\V_{t+1}=\V_t+\frac{\kappa}{2}\x_t\x_t^\top$ for $\kappa>0$, and the initialized matrix $\V_1 = \lambda\mathbf{I}_d$ for $\lambda>0$. Subsequently, OL$^2$M constructs a confidence region $\C_{t+1}$ centered at the estimated parameter $\hat{\bt}_{t+1}^{N}$, such that
\begin{equation}\label{cf-ONS}
\C_{t+1}=\left\{\bt\in\R^d\big\vert\norm{\bt-\hat{\bt}_{t+1}^N}_{\V_{t+1}}^2\leq\gamma_{t+1}\right\}
\end{equation}
where $\gamma_{t+1}=O(d\log t)$ indicating the uncertainty of the estimation and the unknown parameter $\bt_*$ lies in this region with high probability. Finally, OL$^2$M selects the most promising arm $\x_{t+1}$ according to the principle of ``optimization in the face of uncertainty'', i.e.,
\begin{equation}\label{bi-ONS}
(\x_{t+1},\tilde{\bt}_{t+1}) = \argmax_{\x\in\D_{t+1},\bt\in\C_{t+1}}\langle\x,\bt\rangle.
\end{equation}

\subsection{Bandit Learning with Heavy-tailed rewards}
Most of the existing work developed heavy-tailed bandit algorithms using truncation and median of means strategies \citep{Bubeck:2013,Medina:2016,Han:2018,xue:2020,Huang:2022}. \citet{Bubeck:2013} first conducted extensive research on multi-armed bandits with heavy-tailed rewards and achieved a logarithmic regret bound. \citet{Medina:2016} extended it to the SLB model and introduced two algorithms that achieve regret bounds of $\widetilde{O}(dT^{\frac{2+\epsilon}{2(1+\epsilon)}})$ and $\widetilde{O}(d^\frac{1}{2}T^{\frac{1+2\epsilon}{1+3\epsilon}}+dT^{\frac{1+\epsilon}{1+3\epsilon}})$, respectively. \citet{Han:2018} improved upon the results of \citet{Medina:2016} by a more delicate application of heavy-tailed strategies, achieving a regret bound of $\widetilde{O}(dT^{\frac{1}{1+\epsilon}})$. \citet{xue:2020} investigated the case with finite arms and provided two algorithms that attained regret bounds of $\widetilde{O}(d^\frac{1}{2}T^{\frac{1}{1+\epsilon}})$. Recently, \citet{Han:2021} proposed the mean of medians estimator for the super heavy-tailed bandit problem, but the rewards are limited to symmetric distributions. Applying this estimator to the GLB algorithm of \citet{Jun:2017} yields a heavy-tailed GLB algorithm that achieves the regret bound of $O(d(\log T)^{\frac{1}{2\alpha}+\frac{3}{2}}T^{\frac{1}{2}})$ for some $\alpha\in(0,1)$. To illustrate the basic idea of adopting different heavy-tailed strategies in the bandit model, we briefly describe three representative algorithms.

For the algorithm exploiting truncation strategy, we take the algorithm TOFU as an instance \citep{Han:2018}. With the trials up to round $t$, TOFU truncates the rewards $d$ times as follows,
\begin{equation}\label{tm-tofu}
\overline{Y}_t^i=\left[y_1\I_{|u_1^i(t)y_1|\leq h_t},\ldots,y_t\I_{|u_t^i(t)y_t|\leq h_t}\right], i=1,2,\ldots,d
\end{equation}
where $h_t=O(t^{\frac{1-\epsilon}{2(1+\epsilon)}})$ is the truncated criterion, and $u_\tau^i(t)$ denotes the element in the $i$-th row and $\tau$-th column of matrix $\widetilde{\V}_{t+1}^{-1/2}\A_t$, $\A_t=[\x_1,\x_2,\ldots,\x_t]\in\R^{d\times t}$ is the matrix composed of selected arms and $\widetilde{\V}_{t+1}=\A_t\A_t^\top+\mathbf{I}_d$. Using these truncated rewards, TOFU conducts an estimator as $\tilde{\bt}_{t+1}=\widetilde{\V}_{t+1}^{-1/2}[\u_t^1\cdot \overline{Y}_t^1,\ldots,\u_t^d\cdot \overline{Y}_t^d]$ with $\u_t^i\cdot\overline{Y}_t^i=\sum_{\tau=1}^tu_\tau^i(t)y_\tau\I_{|u_\tau^iy_\tau|\leq h_t}$. TOFU then constructs a confidence region centered on this estimator and selects the promising arm, similar to \eqref{cf-ONS} and \eqref{bi-ONS}. Notice that the scalarized parameters $\{u_\tau^i(t)\}_{\tau=1}^t$ are updated at each epoch, requiring TOFU to store the learning history and truncate all rewards at each epoch. Thus, TOFU is not an online method.

For the algorithm exploiting median of means strategy, it's common to play the chosen arm $r$ times and get $r$ sequences of rewards $\{Y_t^j\}_{j=1}^r$, where $Y_t^j=[y_1^j,\ldots,y_t^j]$ is the $j$-th reward sequence up to epoch $t$. MENU executes least square estimation for each reward sequence and get $r$ estimators, i.e.,
\begin{equation}\label{LSE-MM}
\hat{\bt}_{t+1}^{j}=\argmin_{\bt\in\R^d}\norm{\A_t^\top\bt-Y_t^j}_2^2+\norm{\bt}_2^2,\ j=1,2,\ldots,r
\end{equation}
where $r=O(\log T)$ \citep{Han:2018}. Then, the median of means strategy adopted by MENU is operated as follows, 
\begin{equation}\label{mm-menu}
m_j= \textnormal{median of }\left\{\norm{\hat{\bt}_{t+1}^j-\hat{\bt}_{t+1}^s}_{\widetilde{\V}_{t+1}}: s=1,\ldots,r\right\}.
\end{equation}
Then, MENU takes the estimator $\hat{\bt}_{t+1}^{k_*}$ with $k_*=\argmin_{j\in\{1,2,\ldots,r\}}\{m_j\}$ as the center of confidence region. Finally, MENU selects the most promising arm similar to \eqref{bi-ONS}.


For the mean of medians method proposed by \citet{Han:2021}, at each epoch $t$, the agent first plays the selected arm $\bar{r}$ times, with a value of $\bar{r}=O((\log T)^{1/\alpha})$ for some $\alpha\in(0,1)$, and then receives rewards $\{y_t^j\}_{j=1}^{\bar{r}}$ for these plays. Subsequently, the agent randomly divides the rewards into multiple groups, with each group contains $\lceil \bar{r}^\alpha\rceil$ rewards. The agent then takes the median of each group and uses the mean of these medians to update the estimator. Notice that the expectation of the median has a bias to the expected reward other than the symmetric distribution. Thus, mean of medians strategy is limited to symmetric distribution. Another point worth mentioning is that $\bar{r}$ is too large to try sufficient different arms. For example, the agent can only play $\lceil T/\bar{r}\rceil=100$ different arms with $T=10^6$ and $\alpha=0.62$, which is obviously unreasonable\footnotemark.

\footnotetext{$\alpha=0.62$ is nearly optimal for $\epsilon=1$ according to the experiments of \citet{Han:2021}.}

\section{Algorithms}
In this section, we first introduce the generalized linear bandit model and then demonstrate two novel algorithms based on truncation and mean of medians, respectively.

\subsection{Learning Model}
The formal description of the generalized linear bandit model is as follows. In each round $t$, an agent plays an arm $\x_t \in \D_t$ and obtains a stochastic reward $y_t$, which is generated from a generalized linear model represented by the following equation,
\begin{equation}\label{glm}
\Pr(y_t|\x_t)=\exp\left(\frac{y_t\x_t^\top\bt_*-m(\x_t^\top\bt_*)}{g(\tau)}+h(y_t,\tau)\right)
\end{equation}
where $\bt_*$ is the inherent parameters, $\tau>0$ is a known scale parameter, and $ g(\cdot)$ and $h(\cdot, \cdot)$ are normalizers \citep{Book:1989}. The expectation of $y_t$ is given by
\begin{equation*}
\E[y_t|\x_t]=m'(\x_t^\top\bt_*).
\end{equation*}
Thus, $m'(\cdot)$ is the link function in \eqref{expection}, such that  $\mu(\cdot) = m'(\cdot)$. The reward model can be rewritten as
\begin{equation*}\label{model}
y_t=\mu(\x_t^\top\bt_*)+\eta_t
\end{equation*}
where $\eta_t$ is a random noise satisfying the condition
\begin{equation}\label{condition:mean}
\E[\eta_t|\G_{t-1}]=0.
\end{equation}
Here, $\G_{t-1}\triangleq\{\x_1,y_1,\ldots,\x_{t-1},y_{t-1},\x_t\}$ is a $\sigma$-filtration and $\G_0=\emptyset$. Following the existing work \citep{Filippi:2010,Jun:2017,Li:2017}, we make standard assumptions as follows.
\begin{assume}\label{ass:1}
The coefficients $\bt_*$ and contextual vectors $\x$ are bounded, such that $\norm{\bt_*}_2\leq S$ and $\norm{\x}_2\leq 1$ for all $\x\in\D_t$, where $S$ is a known constant.
\end{assume}
\begin{assume}\label{ass:2}
The link function $\mu(\cdot)$ is $L$-Lipschitz on $[-S,S]$, and continuously differentiable on $(-S,S)$. Moreover, there exists some $\kappa>0$ such that $\mu'(z)\geq\kappa$ and $|\mu(z)|\leq U$  for any $z\in(-S,S)$.
\end{assume}

\subsection{Truncation}
Our first algorithm is called Confidence Region with Truncated Mean (CRTM). The complete procedure is provided in Algorithm \ref{alg:CRTM}. Here, we consider the heavy-tailed setting, i.e., there exists a constant $u>0$, the rewards admit
\begin{equation}\label{condition:trunc}
\E\left[|y_t|^{1+\epsilon}|\G_{t-1}\right]\leq u.
\end{equation}

As we have mentioned earlier in Section \ref{related-GLB}, to design effective algorithms for GLB model, constructing a narrow confidence region for the underlying coefficients $\bt_*$ is necessary. However, heavy-tailed rewards that satisfy \eqref{condition:trunc} produce extreme values with high probability, resulting in a confidence region with a large radius. Therefore, a straightforward approach to settle this problem is to truncate the extreme reward to reduce its impact.

\begin{algorithm}[tb]
\caption{Confidence Region with Truncated Mean (CRTM)}
\label{alg:CRTM}
\textbf{Input}: $\delta, \epsilon, u, \kappa, S, \lambda=\max\{1,\kappa/2\}$ and $T\in\N_+$
\begin{algorithmic}[1] 
\State Initialize $\hat{\bt}_1=\bm{0}$ and $\V_1=\lambda\mathbf{I}_d$
\State Define the truncation criterion
$ \tc=2\left(u/\ln(4T/\delta)\right)^\frac{1}{1+\epsilon}\left(d\ln\left(1+\frac{\kappa T}{2\lambda d}\right)/\kappa \right)^\frac{1}{2}T^\frac{1-\epsilon}{2(1+\epsilon)}$
\For {$t=1, 2,\ldots, T$}
\State $(\x_t,\tilde{\bt}_t)=\argmax_{\x\in\D_t,\bt\in\C_{t}}\langle\x,\bt\rangle$
\State Play the arm $\x_t$ and observe the payoff $y_t$
\State Truncate the observed payoff $\tilde{y}_t=y_t\I_{\norm{\x_t}_{\V_t^{-1}}|y_t|\leq \tc}$
\State Compute the gradient $\nabla\tilde{\ell}_t(\hat{\bt}_t)=(-\tilde{y}_t+\mu(\x_t^\top\hat{\bt}_t))\x_t$
\State Update $\V_{t+1}=\V_t+\frac{\kappa}{2}\x_t\x_t^\top$
\State Update the estimator
\begin{equation*}
\hat{\bt}_{t+1}=\argmin_{\norm{\bt}_2\leq S}\frac{\norm{\bt-\hat{\bt}_t}^2_{\V_{t+1}}}{2}+\langle\bt-\hat{\bt}_t,\nabla\tilde{\ell}_t(\hat{\bt}_t)\rangle
\end{equation*}
\State Construct the confidence region
\begin{equation*}
\C_{t+1}=\left\{\bt\in\R^d\big\vert\norm{\bt-\hat{\bt}_{t+1}}^2_{\V_{t+1}}\leq\gamma\right\}
\end{equation*}
\EndFor
\end{algorithmic}
\end{algorithm}

In each round $t$, CRTM first plays an arm $\x_t\in\D_t$ and observes the corresponding reward $y_t$. Then, CRTM truncates the reward $y_t$ using a uniform criterion $\tc = \widetilde{O}(T^{\frac{1-\epsilon}{2(1+\epsilon)}})$, such that
\begin{equation*}
\tilde{y}_t=y_t\I_{\norm{\x_t}_{\V_t^{-1}}|y_t|\leq \tc}
\end{equation*}
where $\V_t=\V_{t-1}+\frac{\kappa}{2}\x_{t-1}\x_{t-1}^\top$ with $\V_1=\lambda \mathbf{I}_d$. Here, $\kappa$ is defined in Assumption \ref{ass:2} and $\lambda=\max\{1, \kappa/2\}$. With the processed action-reward pair $(\x_t,\tilde{y}_t)$, CRTM computes the gradient of the loss function as
\begin{equation}\label{gradient}
\nabla\tilde{\ell}_t(\bt)=(-\tilde{y}_t+\mu(\x_t^\top\bt))\x_t,
\end{equation}
where $\tilde{\ell}_t(\cdot)$ is the negative log-likelihood of the generalized linear model \eqref{glm}. After that, CRTM employs a variant of online Newton step (ONS) to update its estimator, given by
\begin{equation*}\label{opt-learner}
\begin{aligned}
\hat{\bt}_{t+1}=\argmin_{\norm{\bt}_2\leq S}\frac{\norm{\bt-\hat{\bt}_t}^2_{\V_{t+1}}}{2}+\langle\bt-\hat{\bt}_t,\nabla\tilde{\ell}_t(\hat{\bt}_t)\rangle.
\end{aligned}
\end{equation*}
Equipped with above estimation, CRTM constructs the confidence region $\C_{t+1}$ where the inherent parameters $\bt_*$ lies in with high probability, such that
\begin{equation*}\label{confi-reg}
\C_{t+1}=\left\{\bt\in\R^d\big\vert\norm{\bt-\hat{\bt}_{t+1}}^2_{\V_{t+1}}\leq\gamma\right\}
\end{equation*}
where $\gamma=\widetilde{O}(T^\frac{1-\epsilon}{1+\epsilon})$ denotes the width of the confidence region, and details are shown in Theorem~\ref{thm:3}. Given the confidence region $\C_{t+1}$, the most promising arm $\x_{t+1}$ can be obtained through the following maximize operation,
\begin{equation*}\label{equ-opt}
(\x_{t+1},\tilde{\bt}_{t+1})=\argmax_{\x\in\D_{t+1},\bt\in\C_{t+1}}\langle\x,\bt\rangle
\end{equation*}
since $\mu(\cdot)$ is monotonically increasing according to Assumption \ref{ass:2}.

Although there exists several heavy-tailed linear bandit algorithms based on the truncation strategy, such as TOFU \citep{Han:2018} and BTC \citep{xue:2020}, CRTM differs from them in two aspects. Firstly, both TOFU and BTC have to store the historical rewards and truncate them at each epoch, resulting in a computational complexity of $O(T^2)$. In contrast, CRTM achieves online learning by processing only the reward of current round, whose computational complexity is $O(T)$. Secondly, while TOFU and BTC are designed for SLB model and calculate the estimator via least-squares estimation, CRTM is designed for the GLB model and updates the estimator using the ONS method, which makes the analytical techniques fundamentally different. Theorem~\ref{thm:3} provides a tight confidence region, and its proof relies on the induced method because ONS is an iteratively updated method. Due to the page limit, we provide the detailed proof in the Appendix A.2.

\begin{thm}\label{thm:3}
If the rewards satisfy \eqref{condition:mean} and \eqref{condition:trunc}, then with probability as least $1-\delta$, the confidence region in CRTM is 
\begin{equation*}
\norm{\bt-\hat{\bt}_{t+1}}_{\V_{t+1}}^2\leq\gamma, \forall t\geq0
\end{equation*}
where 
\begin{equation*}
\begin{aligned}
\gamma=224u^\frac{2}{1+\epsilon}\ln(4T/\delta)^\frac{2\epsilon}{1+\epsilon}T^\frac{1-\epsilon}{1+\epsilon}\frac{4d}{\kappa}\ln\left(1+\frac{\kappa T}{2\lambda d}\right)+2\lambda S^2+\frac{48U^2d}{\kappa}\ln\left(1+\frac{\kappa T}{2\lambda d}\right).
\end{aligned}
\end{equation*}
\end{thm}

With above confidence region, the regret bound of CRTM is explicitly given as follows.

\begin{thm}\label{thm:4}
If the rewards satisfy \eqref{condition:mean} and \eqref{condition:trunc}, then with probability at least $1-\delta$, the regret of CRTM satisfies
\begin{equation*}
R(T)\leq O\left(d(\log T)^\frac{1+2\epsilon}{1+\epsilon}T^{\frac{1}{1+\epsilon}}\right).
\end{equation*}
\end{thm}

\noindent\textbf{Remark:} The above theorem establishes a $\widetilde{O}(dT^{\frac{1}{1+\epsilon}})$ regret bound with the assumption that the $(1+\epsilon)$-th moment of the rewards is bounded for some $\epsilon\in(0,1]$. Existing algorithms based on truncation is time-consuming because they need to store the learning history and truncate all historical rewards at each epoch \citep{Han:2018,xue:2020}. Unlike the recently proposed mean of medians method which is limited in symmetric rewards \citep{Han:2021}, CRTM expands it to asymmetric and achieves an improved regret bound by a factor of $O((\log T)^\frac{1}{2\alpha})$ for some $\alpha\in(0,1)$ if $\epsilon = 1$. Furthermore, CRTM is almost optimal as the lower bound is $\Omega(dT^{\frac{1}{1+\epsilon}})$ \citep{Han:2018}.

\subsection{Mean of Medians}
In this section, we present our second algorithm, referred to as Confidence Region with Mean of Medians (CRMM), which shares a similar framework with CRTM but uses a different mean of medians estimator. The complete procedure is outlined in Algorithm \ref{alg:CRMM}. CRMM requires that for some $\epsilon\in(0,1]$, the $1+\epsilon$ central moment of the rewards is bounded, and the distribution of rewards is symmetric. Precisely, for some $\epsilon\in(0,1]$, there exists a constant $v>0$ such that the rewards satisfy
\begin{equation}\label{condition:mom}
\E\left[|\eta_t|^{1+\epsilon}|\G_{t-1}\right]\leq v\  \text{and}\ p(\eta_t)=p(-\eta_t).
\end{equation}

\begin{algorithm}[tb]
\caption{Confidence Region with Mean of Medians (CRMM)}
\label{alg:CRMM}
\textbf{Input}: $\delta, \epsilon, v, \kappa, S, \lambda=\max\{1,\kappa/2\}$ and $T\in\N_+$
\begin{algorithmic}[1] 
\State Initialize $\hat{\bt}_1=\bm{0}, \V_1=\lambda\mathbf{I}_d$ and $\gamma_1=\lambda S^2$
\State $r=\left \lceil 16\ln\frac{4T}{\delta} \right \rceil$ and $T_0=\lfloor T/r\rfloor$
\For {$t=1, 2,\ldots, T_0$}
\State $(\x_t,\tilde{\bt}_t)=\argmax_{\x\in\D,\bt\in\C_{t}}\langle\x,\bt\rangle$
\State Play the arm $\x_t$ $r$ times and observe the rewards $\{y_t^1,y_t^2,\ldots,y_t^r\}$
\State Take the median of $\{y_t^1,y_t^2,\ldots,y_t^r\}$ as $\bar{y}_t$
\State Compute the gradient $\nabla\bar{\ell}_t(\hat{\bt}_t)=(-\bar{y}_t+\mu(\x_t^\top\hat{\bt}_t))\x_t$
\State Update $\V_{t+1}=\V_t+\frac{\kappa}{2}\x_t\x_t^\top$
\State Compute the center of confidence region
\begin{equation*}
\hat{\bt}_{t+1}=\argmin_{\norm{\bt}_2\leq S}\frac{\norm{\bt-\hat{\bt}_t}^2_{\V_{t+1}}}{2}+\langle\bt-\hat{\bt}_t,\nabla\bar{\ell}_t(\hat{\bt}_t)\rangle
\end{equation*}
\State Construct the confidence region
\begin{equation*}
\C_{t+1}=\left\{\bt\in\R^d\big\vert\norm{\bt-\hat{\bt}_{t+1}}^2_{\V_{t+1}}\leq\gamma_{t+1}\right\}
\end{equation*}
\EndFor
\end{algorithmic}
\end{algorithm}

At each epoch $t$, CRMM plays the selected arm $\x_t$ $r$ times, generating rewards $\{y_t^1,\ldots,y_t^r\}$ with $r=O(\log T)$. To obtain a robust estimation using mean of medians strategy, CRMM first takes the median of $\{y_t^1,\ldots,y_t^r\}$, denoted by $\bar{y}_t$. Subsequently, CRMM computes the gradient with the arm-reward pair $(\x_t, \bar{y}_t)$ through the operation similar to \eqref{gradient}. Then, CRMM employs a variant of ONS to update the estimator and construct the confidence region $\C_{t+1}$ centered on the new estimator. The details about the constructed confidence region is given in Theorem~\ref{thm:1}.

Compared to existing bandit algorithms that utilize the median of means strategy, the primary difference lies in the item chosen as the ``means''. As we have introduced in \eqref{mm-menu}, MENU of \citet{Han:2018} uses the distance between different estimators as the ``means''. BMM of \citet{xue:2020} calculates multiple estimated rewards for each arm and treats them as the ``means''. Both MENU and BMM require $O(\log T)$ estimators during each round, whereas CRMM only requires one estimator. Moreover, compared to the mean of medians approach \citep{Han:2021}, CRMM plays each selected arm fewer times, leading to more model updates, which is critical based on experimental results. Since the chosen arm has to be played multiple times, we assume the arm set for CRMM is static, such that $\D_t=\D$ for $t>0$, which is a common assumption \citep{Medina:2016,Zhang:2016,Lu:2019-2}. The following theorem guarantees a tight confidence region.

\begin{thm}\label{thm:1}
If the rewards satisfy \eqref{condition:mean} and \eqref{condition:mom}, then with probability as least $1-2\delta$, the confidence region in CRMM is 
\begin{equation*}
\norm{\bt-\hat{\bt}_{t+1}}^2_{\V_{t+1}}\leq\gamma_{t+1}, \forall t\geq0
\end{equation*}
where 
\begin{equation*}
\begin{aligned}
\gamma_{t+1}=&\left(4U^2+C\rho t^{\frac{1-\epsilon}{1+\epsilon}}\right)\frac{4d}{\kappa}\ln\left(1+\frac{\kappa t}{2\lambda d}\right)+\lambda S^2+\frac{2\rho^2}{\kappa} t^{\frac{1-\epsilon}{1+\epsilon}},\\
\rho=&2C\ln(4T/\delta)+2C^{-\epsilon}rv,\ C=(4v)^{\frac{1}{1+\epsilon}}.
\end{aligned}
\end{equation*}
\end{thm}
With above confidence region, we prove the following regret bound for CRMM.

\begin{thm}\label{thm:2}
If the rewards satisfy \eqref{condition:mean} and \eqref{condition:mom}, then with probability at least $1-2\delta$, the regret of CRMM satisfies
\begin{equation*}
R(T)\leq O\left(d(\log T)^{\frac{3}{2}+\frac{\epsilon}{1+\epsilon}}T^{\frac{1}{1+\epsilon}}\right).
\end{equation*}
\end{thm}

\noindent\textbf{Remark:} Theorem~\ref{thm:1} clarifies that if the rewards have a finite $1+\epsilon$ central moment for some $\epsilon\in(0,1]$, CRMM achieves a regret bound of $\widetilde{O}(dT^{\frac{1}{1+\epsilon}})$. This bound reduces to $\widetilde{O}(d\sqrt{T})$ when $\epsilon=1$, indicating that CRMM achieves the same order as the bounded rewards assumption regarding both $d$ and $T$ \citep{Zhang:2016,Jun:2017}. Compared to the approach of \citet{Han:2021}, CRMM enhances the bound by an order of $O((\log T)^{\frac{1}{2\alpha}-\frac{1}{2}})$ for a fixed $\alpha\in(0,1)$ if $\epsilon = 1$.

\section{Experiments}\label{experiments}

This section demonstrates the improvement of our algorithms by numerical experiments. Firstly, we show the effectiveness of our algorithms in dealing with heavy-tailed problems by comparing their regret to that of existing generalized linear bandit algorithms. Secondly, we evaluate the efficiency of our algorithms by comparing their time consumption to other existing algorithms designed for heavy-tailed bandit problems. All algorithms are implemented using PyCharm 2022 and tested on a laptop with a 2.5GHz CPU and 32GB of memory.

\subsection{Regret Comparison}
To assess the enhancement of our algorithms in handling heavy-tailed problems, we utilize the vanilla GLB algorithms, specifically OL$^2$M \citep{Zhang:2016} and GLOC \citep{Jun:2017}, as baselines. Additionally, we incorporate the mean of medians method proposed by \citet{Han:2021} into OL$^2$M and GLOC, resulting in another two baselines OL$^2$M\_mom and GLOC\_mom, respectively. All algorithms are configured with $\epsilon=1$, $\delta=0.01$, and $T=10^6$.

\begin{figure*}[tb]
\centering 
\subfigure[Student's $t$-noises (symmetric)]{
\includegraphics[width=0.49\textwidth]{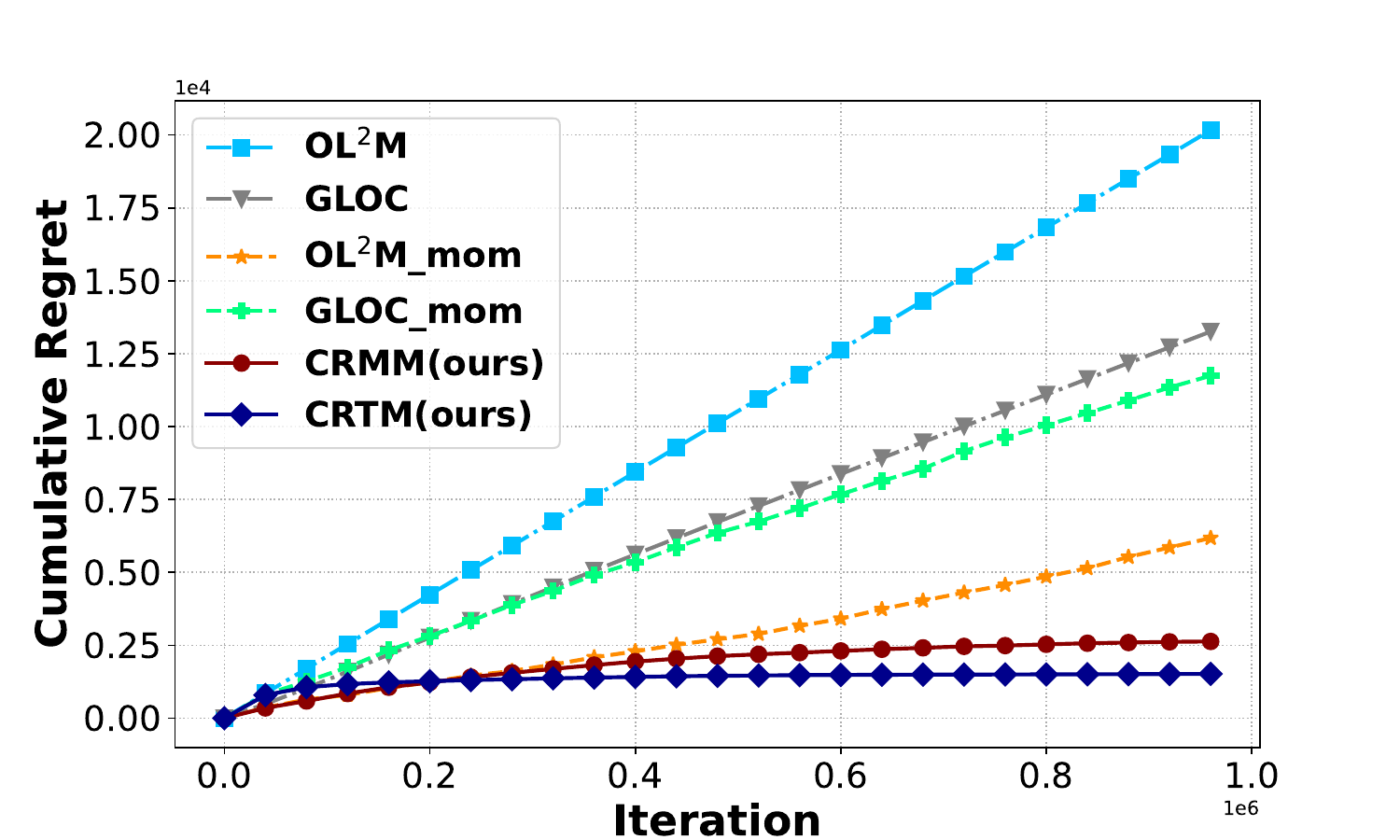}}
\subfigure[Pareto noises (asymmetric)]{
\includegraphics[width=0.49\textwidth]{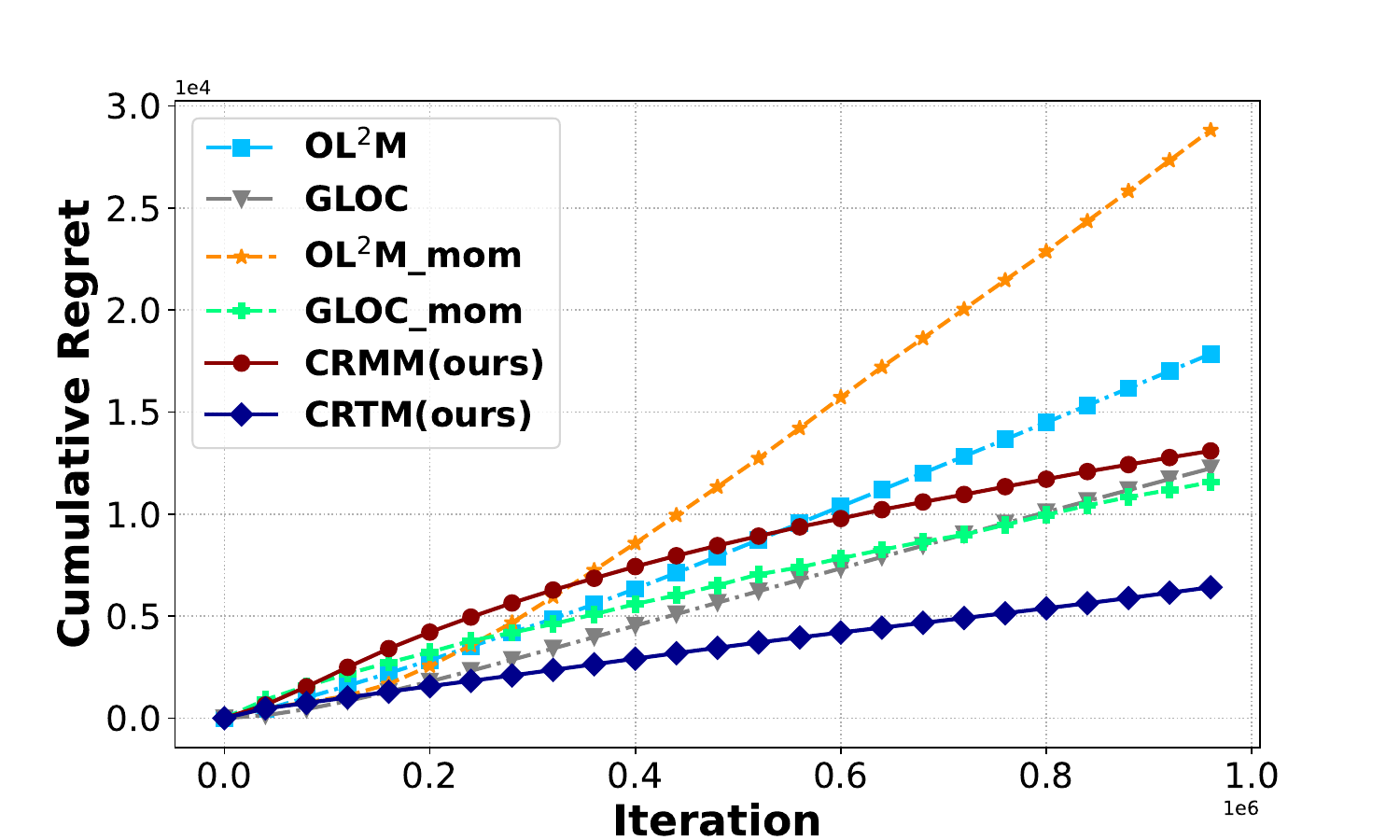}}
\caption{Regret comparison}
\label{Fig-2016-s}
\end{figure*}

Let $\bt_*=\ones/\sqrt{d}\in\R^d$, where $\ones$ is an all-$1$ vector and $\norm{\bt_*}_2=1$. The number of arms is set to $K=20$, and the feature dimension is $d=10$. Each component of the contextual vector $\x_t$ is uniformly sampled from the interval $[0,1]$, and then normalized to be unit norm, i.e., $\norm{\x_t}_2=1$. We tune the width of the confidence region following the common practice in bandit learning \citep{Zhang:2016,Jun:2017}. Precisely, with $c$ being a tuning parameter searched within $[1e^{-4},1]$, the width of the confidence region for OL$^2$M and GLOC are set as $\gamma_t=cd\ln(t/\lambda+1)$ and $\gamma_t=c\sum_{\tau=1}^t(\mu(\x_\tau^\top\hat{\bt}_\tau)-y_\tau)^2\norm{\x_\tau}_{\V_\tau^{-1}}^2$, respectively. In addition, the radius of the confidence region is set as $cd\ln(4T/\delta)^\frac{2\epsilon}{1+\epsilon}\ln(T/(d\lambda)+1)T^\frac{1-\epsilon}{1+\epsilon}$ for CRTM, and $cd\ln(t/(d\lambda)+1)t^\frac{1-\epsilon}{1+\epsilon}$ for CRMM. For OL$^2$M\_mom and GLOC\_mom, the chosen arm is played $\bar{r}=(16\ln(2T/\delta))^{1/\alpha}$ times per round, and $\alpha=0.62$ is close to optimal according to the experiments of \cite{Han:2021}.

We run 10 repetitions for each algorithm and display the average regret with time evolution. According to the generalized linear bandit model, the observed reward $y_t$ is given by
\begin{equation*}
y_t=\mu(\x_t^\top\bt_*)+\eta_t
\end{equation*}
where $\mu(x)=\frac{1}{1+e^{-x}}$ is the logit model and $\eta_t$ is the heavy-tailed noises. To evaluate the algorithms performance under both symmetric and asymmetric rewards, $\eta_t$ fits the following two distributions,
\begin{enumerate}[(i)]
\item Student's $t$-Noise: $\eta_t\sim\frac{G(2)}{\sqrt{3\pi}G(1.5)}\left(1+\frac{x^2}{3}\right)^{-2}$ where $G(\cdot)$ is the Gamma function;

\item Pareto Noise: $\eta_t\sim\frac{sx_m^s}{x^{s+1}}\mathbb{I}_{x\geq x_m}$ where $s=3$ and $x_m=0.01$.
\end{enumerate}

Fig.~\ref{Fig-2016-s} compares our algorithms against two vanilla GLB algorithms (OL$^2$M and GLOC), as well as these two algorithms exploiting mean of medians estimators (OL$^2$M\_mom and GLOC\_mom). Fig.~\ref{Fig-2016-s}(a) shows that CRTM and CRMM outperform the other four algorithms. CRTM provides the best performance, which is consistent with the theoretical guarantees. OL$^2$M\_mom and GLOC\_mom appear ineffective at handling heavy-tailed problems, because they update estimator only $100$ times with the chosen arm played $\bar{r}$ times \citep{Han:2021}. Fig.~\ref{Fig-2016-s}(b) presents the cumulative regrets under asymmetric noises, with CRTM still having the lowest regret curve, demonstrating its generality and effectiveness in handling heavy-tailed bandit problems. On the other hand, CRMM, GLOC\_mom, and OL$^2$M\_mom performs poorly in Fig.~\ref{Fig-2016-s}(b), as they can not deal with the asymmetric rewards. 

\subsection{Runtime Comparison}

To demonstrate the efficiency improvement of our algorithms, we compare them with existing heavy-tailed bandit algorithms such as CRT and MoM \citep{Medina:2016}, TOFU and MENU \citep{Han:2018}, and SupBTC and SupBMM \citep{xue:2020}. Among them, CRT, TOFU and SupBTC employ truncation strategy, while MoM, MENU and SupBMM utilize the median of means strategy. 

\begin{wraptable}{r}{7.3cm}
\caption{Runtime comparsion}
\centering
\begin{tabular}{l l l l }
\hline
 Algorithm & Time(s) & Algorithm & Time(s)\\
\hline
CRT & 3.1737 & MoM & 0.0630 \\
TOFU & 3931.9963& MENU & 24.1990\\
SupBTC & 1187.1863& SupBMM & 0.0685 \\
CRTM & \textbf{2.2909} & CRMM & \textbf{0.0514} \\
\hline
\end{tabular}
\label{table2}
\end{wraptable}

The experimental settings are the same as described in Regret Comparison section, except for the time horizon and feature dimension. We use a smaller time horizon $T=10^4$ since TOFU is time-consuming. The feature dimension is increased to $d = 100$ to highlight the difference between SupBTC and TOFU. The computational runtimes are shown in Table~\ref{table2}.


For the truncation-based algorithms, CRTM consumes the least time, while TOFU and SupBTC takes over a hundred times longer to execute than CRTM, representing a significant improvement. CRT takes only slightly longer than CRTM as both algorithms update the model online, but the regret bound of CRT is $\widetilde{O}(dT^\frac{3}{4})$, which is $\widetilde{O}(T^\frac{1}{4})$ worse than the bound of CRTM. For median of means algorithms, CRMM has the shortest runtime. MENU takes significantly longer than the other algorithms because MENU needs to calculate the distance between $O(\log T)$ estimators.

\section{Conclusion and Future Work}

We present two algorithms, CRTM and CRMM, for the generalized linear bandit model with heavy-tailed rewards, which utilize the truncation and mean of medians strategies, respectively. Both algorithms achieve the regret bound of $\widetilde{O}(dT^{\frac{1}{1+\epsilon}})$ conditioned on a bounded $(1+\epsilon)$-th moment of rewards, where $\epsilon\in(0,1]$. This bound is almost optimal since the lower bound of the stochastic liear bandit problem is $\Omega(dT^{\frac{1}{1+\epsilon}})$ \citep{Han:2018}. CRTM is the first truncation-based online algorithm for the heavy-tailed bandit problem that handles both symmetric and asymmetric rewards and approaches the optimal regret bound. CRMM enhances the regret bound of the the most related work by a logarithmic factor \citep{Han:2021}. However, CRMM is limited to symmetric rewards and we will investigate to overcome this restriction in the future.

\begin{ack}
This work was partially supported by the National Key R\&D Program of China~(2022ZD0114801), the Key Basic Research Foundation of Shenzhen~(JCYJ20220818100005011), NSFC~(62122037, 61921006) and the major key project of PCL~(PCL2021A12).
\end{ack}

\bibliographystyle{named}
\bibliography{ref}

\newpage
\appendix
\rule[-10pt]{5.5in}{4pt}
\begin{center}
{\Large \bfseries Supplementary Material: Efficient Algorithms for Generalized Linear Bandits with Heavy-tailed Rewards}
\end{center}
\rule[-1pt]{5.5in}{1pt}

\section{Proof of Theorem \ref{thm:3}}


To ensure clarity of expression, we have divided the proof of Theorem \ref{thm:3} into two subsections. The first subsection establishes a general upper bound for the confidence region constructed by ONS. Building upon the first subsection, we employ the truncated technique in the second subsection to deduce the confidence region for CRTM.

\subsection {General Upper Bound of ONS}

For the sake of representation, we define the loss function for the action-reward pair $(\x_t, y_t)$ as
\begin{equation*}
\ell_t(\bt)=-y_t\x_t^\top\bt+m(\x_t^\top\bt),
\end{equation*}
and the conditional expectation for this loss function is denoted as $f_t(\bt)=\E[\ell_t(\bt)|\G_{t-1}]$. 

First, we propose the following lemma to display the strong convexity of the loss function.

\begin{lem}\label{lem:1}
For any $\bt_1,\bt_2\in\R^d$ satisfying $\norm{\bt_1}_2\leq S, \norm{\bt_2}_2\leq S$, the inequality 
\begin{equation*}
\begin{aligned}
\ell_t(\bt_1)-\ell_t(\bt_2)\geq\nabla\ell_t(\bt_2)^\top(\bt_1-\bt_2)+\frac{\kappa}{2}\left(\x_t^\top\bt_1-\x_t^\top\bt_2\right)^2
\end{aligned}
\end{equation*}
is true for all $t>0$.
\end{lem}

\textbf{Proof.} Let $L_t(z)=-y_tz+m(z), z\in[-S,S]$, then $L_t''(z)=\mu'(z)\geq\kappa$ due to Assumption~\ref{ass:2}. Thus, $L_t(z)$ is a $\kappa$-strongly convex function, which indicates that 
\begin{equation*}
L_t(z_1)-L_t(z_2)\geq L'_t(z_2)(z_1-z_2)+\frac{\kappa}{2}\left(z_1-z_2\right)^2.
\end{equation*}
Let $z_1=\x_t^\top\bt_1$ and $z_2=\x_t^\top\bt_2$, we get that
\begin{equation*}\label{lem1:ineq-1}
\begin{aligned}
L_t(\x_t^\top\bt_1)-L_t(\x_t^\top\bt_2)\geq& L_t'(\x_t^\top\bt_2)(\x_t^\top\bt_1-\x_t^\top\bt_2)+\frac{\kappa}{2}\left(\x_t^\top\bt_1-\x_t^\top\bt_2\right)^2.
\end{aligned}
\end{equation*}
Taking $L_t(\x_t^\top\bt)=\ell_t(\bt)$ and $L_t'(\x_t^\top\bt)\x_t=\nabla\ell_t(\bt)$ into above equation finishes the proof.
$\hfill\square$

Then, we propose Lemma~\ref{lem:2} to show that $\bt_*$ is the minimum point of the expected loss function.

\begin{lem}\label{lem:2}
Suppose $\bt\in\R^d$ satisfies $\norm{\bt}_2\leq S$, then $f_t(\bt)-f_t(\bt_*)\geq 0$ for all $t>0$.
\end{lem}

\textbf{Proof.} Recall that GLB model satisfys $\E[y_t|\x_t]=m'(\x_t^\top\bt_*)$ and $\mu(\cdot) = m'(\cdot)$, thus
\begin{align*}
f_t(\bt)-f_t(\bt_*)=~&\E[\ell_t(\bt)-\ell_t(\bt_*)|\G_{t-1}]\\
=~&m(\x_t^\top\bt)-m(\x_t^\top\bt_*)-\mu(\x_t^\top\bt_*)(\x_t^\top\bt-\x_t^\top\bt_*)\\
\geq~&m'(\x_t^\top\bt_*)(\x_t^\top\bt-\x_t^\top\bt_*)-\mu(\x_t\bt_*)(\x_y^\top\bt-\x_t^\top\bt_*)\\
=~&0
\end{align*}
where the inequality holds because $m(\cdot)$ is $\kappa$-strongly convex.
$\hfill\square$

To exploit the property of ONS, we adopt the following lemma from \cite{Zhang:2016}.

\begin{lem}\label{lem:3}
For any $t>0$, the inequality 
\begin{equation*}
\begin{aligned}
\nabla\ell_t(\hat{\bt}_t)^\top(\hat{\bt}_t-\bt_*)-\frac{1}{2}\norm{\nabla\ell_t(\hat{\bt}_t)}_{\V_{t+1}^{-1}}^2\leq~&\frac{1}{2}\left(\norm{\hat{\bt}_t-\bt_*}_{\V_{t+1}}^2-\norm{\hat{\bt}_{t+1}-\bt_*}_{\V_{t+1}}^2\right)
\end{aligned}
\end{equation*}
holds.
\end{lem}
With above three lemmas, we are ready to bound the confience region of the ONS estimation. Lemma~\ref{lem:1} tells that
\begin{equation*}
\begin{aligned}
\ell_t(\hat{\bt}_t)-\ell_t(\bt_*)\leq~\nabla\ell_t(\hat{\bt}_t)^\top(\hat{\bt}_t-\bt_*)-\frac{\kappa}{2}\left(\x_t^\top\hat{\bt}_t-\x_t^\top\bt_*\right)^2.
\end{aligned}
\end{equation*}
If we take expectation in both sides, it becomes
\begin{align*}
f_t(\hat{\bt}_t)-f_t(\bt_*)\leq~\nabla f_t(\hat{\bt}_t)^\top(\hat{\bt}_t-\bt_*)-\frac{\kappa}{2}\left(\x_t^\top\hat{\bt}_t-\x_t^\top\bt_*\right)^2.
\end{align*}
Lemma~\ref{lem:2} tells that
\begin{equation}\label{ineq:0519-6}
\begin{aligned}
0\leq~&\nabla f_t(\hat{\bt}_t)^\top\left(\hat{\bt}_t-\bt_*\right)-\frac{\kappa}{2}\left(\x_t^\top\hat{\bt}_t-\x_t^\top\bt_*\right)^2\\
=~&\left(\nabla f_t(\hat{\bt}_t)-\nabla \ell_t(\hat{\bt}_t)\right)^\top\left(\hat{\bt}_t-\bt_*\right)-\frac{\kappa}{2}\left(\x_t^\top\hat{\bt}_t-\x_t^\top\bt_*\right)^2+\nabla \ell_t(\hat{\bt}_t)^\top(\hat{\bt}_t-\bt_*).
\end{aligned}
\end{equation}
According to Lemma~\ref{lem:3}, we can relax the last term in the right side of \eqref{ineq:0519-6} and get
\begin{equation}\label{0111-6}
\begin{aligned}
0\leq~&\left(\nabla f_t(\hat{\bt}_t)-\nabla \ell_t(\hat{\bt}_t)\right)^\top\left(\hat{\bt}_t-\bt_*\right)-\frac{\kappa}{2}\left(\x_t^\top\hat{\bt}_t-\x_t^\top\bt_*\right)^2\\
&+\frac{1}{2}\norm{\nabla\ell_t(\hat{\bt}_t)}_{\V_{t+1}^{-1}}^2+\frac{1}{2}\left(\norm{\hat{\bt}_t-\bt_*}_{\V_{t+1}}^2-\norm{\hat{\bt}_{t+1}-\bt_*}_{\V_{t+1}}^2\right).
\end{aligned}
\end{equation}
Then, taking the gradient
\begin{align*}
\nabla \ell_t(\hat{\bt}_t)=-y_t\x_t+\mu(\x_t^\top\hat{\bt}_t)\x_t,~ \nabla f_t(\hat{\bt}_t)=-\mu(\x_t^\top\bt_*)\x_t+\mu(\x_t^\top\hat{\bt}_t)\x_t
\end{align*}
into inequality \eqref{0111-6}, we get that
\begin{equation*}
\begin{aligned}
0\leq~&\frac{1}{2}\left(\norm{\hat{\bt}_t-\bt_*}_{\V_{t+1}}^2-\norm{\hat{\bt}_{t+1}-\bt_*}_{\V_{t+1}}^2\right)-\frac{\kappa}{2}\left(\x_t^\top\hat{\bt}_t-\x_t^\top\bt_*\right)^2 \\
&+\left(y_t-\mu(\x_t^\top\bt_*)\right)\x_t^\top(\hat{\bt}_t-\bt_*)+\frac{1}{2}\norm{(-y_t+\mu(\x_t^\top\hat{\bt}_t))\x_t}_{\V_{t+1}^{-1}}^2.
\end{aligned}
\end{equation*}
A simple application of triangle inequality tells that 

\begin{equation*}
\begin{aligned}
0\leq~&\frac{1}{2}\left(\norm{\hat{\bt}_t-\bt_*}_{\V_{t+1}}^2-\norm{\hat{\bt}_{t+1}-\bt_*}_{\V_{t+1}}^2\right)-\frac{\kappa}{2}\left(\x_t^\top\hat{\bt}_t-\x_t^\top\bt_*\right)^2\\
&+\left(y_t-\mu(\x_t^\top\bt_*)\right)\x_t^\top(\hat{\bt}_t-\bt_*)\\
&+\frac{1}{2}(y_t-\mu(\x_t^\top\bt_*))^2\norm{\x_t}_{\V_{t+1}^{-1}}^2+\frac{1}{2}(\mu(\x_t^\top\bt_*)-\mu(\x_t^\top\hat{\bt}_t))^2\norm{\x_t}_{\V_{t+1}^{-1}}^2\\
\leq~&\frac{1}{2}\left(\norm{\hat{\bt}_t-\bt_*}_{\V_t}^2-\norm{\hat{\bt}_{t+1}-\bt_*}_{\V_{t+1}}^2\right)-\frac{\kappa}{2}\left(\x_t^\top\hat{\bt}_t-\x_t^\top\bt_*\right)^2\\
&+\left(y_t-\mu(\x_t^\top\bt_*)\right)\x_t^\top(\hat{\bt}_t-\bt_*)\\
&+\frac{1}{2}(y_t-\mu(\x_t^\top\bt_*))^2\norm{\x_t}_{\V_t^{-1}}^2+\frac{1}{2}(\mu(\x_t^\top\bt_*)-\mu(\x_t^\top\hat{\bt}_t))^2\norm{\x_t}_{\V_t^{-1}}^2
\end{aligned}
\end{equation*}
where the second equality holds because $\V_{t+1}=\V_t+\frac{\kappa}{2}\x_t\x_t^\top$. By summing the above inequality from $1$ to $t$ and rearranging, the confidence region can be bounded as
\begin{equation}\label{ineq:25}
\begin{aligned}
&\norm{\hat{\bt}_{t+1}-\bt_*}_{\V_{t+1}}^2\\
\leq~& \norm{\hat{\bt}_1-\bt_*}_{\V_1}^2-\frac{\kappa}{2}\sum_{\tau=1}^t\left(\x_\tau^\top\hat{\bt}_\tau-\x_\tau^\top\bt_*\right)^2+\sum_{\tau=1}^t\left(\mu(\x_\tau^\top\bt_*)-\mu(\x_\tau^\top\hat{\bt}_\tau)\right)^2\norm{\x_\tau}_{\V_\tau^{-1}}^2\\
&+\sum_{\tau=1}^t 2\left(y_\tau-\mu(\x_\tau^\top\bt_*)\right)\x_\tau^\top(\hat{\bt}_\tau-\bt_*)+\sum_{\tau=1}^t \left(y_\tau-\mu(\x_\tau^\top\bt_*)\right)^2\norm{\x_\tau}_{\V_\tau^{-1}}^2.
\end{aligned}
\end{equation}

Until now, we have proven an upper bound for the ONS method updated with a general action-reward pair $(\x_t, y_t)$, and the bound is shown in equation~\eqref{ineq:25}. 

\subsection{Truncated Upper Bound of CRTM}

CRTM updates the estimator with a truncated action-reward pair $(\x_t, \tilde{y}_t)$, where $\tilde{y}_t$ is the truncated reward $y_t\I_{\norm{\x_t}_{\V_t^{-1}}|y_t|\leq \tc}$. Replacing the $(\x_t, y_t)$ of general upper bound \eqref{ineq:25} by $(\x_t, \tilde{y}_t)$, we get that
\begin{equation}\label{ineq:28}
\begin{aligned}
&\norm{\hat{\bt}_{t+1}-\bt_*}_{\V_{t+1}}^2\\
\leq~& \norm{\hat{\bt}_1-\bt_*}_{\V_1}^2-\frac{\kappa}{2}\sum_{\tau=1}^t\left(\x_\tau^\top\hat{\bt}_\tau-\x_\tau^\top\bt_*\right)^2+\sum_{\tau=1}^t\left(\mu(\x_\tau^\top\bt_*)-\mu(\x_\tau^\top\hat{\bt}_\tau)\right)^2\norm{\x_\tau}_{\V_\tau^{-1}}^2\\
 &+\sum_{\tau=1}^t 2\left(\tilde{y}_\tau-\mu(\x_\tau^\top\bt_*)\right)\x_\tau^\top\left(\hat{\bt}_\tau-\bt_*\right)+\sum_{\tau=1}^t \left(\tilde{y}_\tau-\mu(\x_\tau^\top\bt_*)\right)^2\norm{\x_\tau}_{\V_\tau^{-1}}^2.
\end{aligned}
\end{equation}
Assumption \ref{ass:2} shows that the upper bound of $\mu(\cdot)$ is $U$. Thus, the inequality \eqref{ineq:28} can be simplified as
\begin{equation*}
\begin{aligned}
\norm{\hat{\bt}_{t+1}-\bt_*}_{\V_{t+1}}^2\leq~& \norm{\hat{\bt}_1-\bt_*}_{\V_1}^2+6U^2\sum_{\tau=1}^t\norm{\x_\tau}_{\V_\tau^{-1}}^2\\
&+2\sum_{\tau=1}^t\left(\tilde{y}_\tau-\mu(\x_\tau^\top\bt_*)\right)\x_\tau^\top\left(\hat{\bt}_\tau-\bt_*\right)\\
&+2\sum_{\tau=1}^t \norm{\x_\tau}_{\V_\tau^{-1}}^2y_\tau^2\I_{\norm{\x_\tau}_{\V_\tau^{-1}}|y_\tau|\leq \tc}.
\end{aligned}
\end{equation*}
We define $\beta_\tau=\norm{\x_\tau}_{\V_\tau^{-1}}$. Since $\V_1=\lambda\mathbf{I}_d$ and $\hat{\bt}_1=\bm{0}$, we can deduce that
\begin{equation}\label{ineq:0101-19}
\begin{aligned}
\norm{\hat{\bt}_{t+1}-\bt_*}_{\V_{t+1}}^2\leq~& \lambda S^2+6U^2\sum_{\tau=1}^t \beta_\tau^2+2\underbrace{\sum_{\tau=1}^t \beta_\tau^2y_\tau^2\I_{|\beta_\tau y_\tau|\leq \tc}}_{A}\\
 &+2\sum_{\tau=1}^t \underbrace{\left(y_\tau\I_{|\beta_\tau y_\tau|\leq \tc}-\mu(\x_\tau^\top\bt_*)\right)\x_\tau^\top\left(\hat{\bt}_\tau-\bt_*\right)}_{B_\tau}.
\end{aligned}
\end{equation}

Then, we will employ analytic techniques of truncated strategy to bound the terms $A$ and $\sum_{\tau=1}^tB_\tau$. 

\begin{lem}\label{0519-1}
Suppose that $\E\left[|y_\tau|^{1+\epsilon}|\G_{\tau-1}\right]\leq u$ for $\tau=1,2,\ldots,t$. Then, we have that
\begin{equation*} \label{new-1015-1}
A\leq2\tc^2\ln(2/\delta)+\frac{3}{2}\tc^{1-\epsilon}\sum_{\tau=1}^t \beta_\tau^{1+\epsilon}u
\end{equation*}
holds with probability at least $1-\delta$.
\end{lem}

\textbf{Proof.} According to the triangle inequality, $A$ can be relaxed as 
\begin{equation}\label{0111-15}
\begin{aligned}
A\leq \left|\sum_{\tau=1}^t \beta_\tau^2y_\tau^2\I_{|\beta_\tau y_\tau|\leq \tc}-\E\left[\beta_\tau^2y_\tau^2\I_{|\beta_\tau y_\tau|\leq \tc}|\G_{\tau-1}\right]\right|+\sum_{\tau=1}^t\E\left[\beta_\tau^2y_\tau^2\I_{|\beta_\tau y_\tau|\leq \tc}|\G_{\tau-1}\right].
\end{aligned}
\end{equation}
In light of Bernstein's inequality \citep[Lemma~11]{Bernstein}, we have that
\begin{equation}\label{0111-16}
\begin{aligned}
&\left|\sum_{\tau=1}^t \beta_\tau^2y_\tau^2\I_{|\beta_\tau y_\tau|\leq \tc}-\E\left[\beta_\tau^2y_\tau^2\I_{|\beta_\tau y_\tau|\leq \tc}|\G_{\tau-1}\right]\right|\\
\leq~&2\tc^2\ln(2/\delta)+\frac{1}{2\tc^2}\sum_{\tau=1}^t \Var\left[\beta_\tau^2y_\tau^2\I_{|\beta_\tau y_\tau|\leq \tc}|\G_{\tau-1}\right]\\
\leq~&2\tc^2\ln(2/\delta)+\frac{1}{2\tc^2}\sum_{\tau=1}^t \beta_\tau^{1+\epsilon}u\tc^{3-\epsilon}
\end{aligned}
\end{equation}
holds with probability at least $1-\delta$, and the second inequality of above equation holds because the $(1+\epsilon)$-th moment of rewards is bounded by $u$.

We can bound the second term in the right side of \eqref{0111-15} as 
\begin{equation}\label{0111-17}
\begin{aligned}
\sum_{\tau=1}^t\E\left[\beta_\tau^2y_\tau^2\I_{|\beta_\tau y_\tau|\leq \tc}|\G_{\tau-1}\right]\leq\tc^{1-\epsilon}\sum_{\tau=1}^t\beta_\tau^{1+\epsilon}u.
\end{aligned}
\end{equation}
Combining the inequalities \eqref{0111-15}, \eqref{0111-16} and \eqref{0111-17} finishes the proof of Lemma~\ref{0519-1}.
$\hfill\square$

We will now proceed to bound the term $\sum_{\tau=1}^tB_\tau$.
\begin{lem}\label{0519-2}
Suppose that $\E\left[|y_\tau|^{1+\epsilon}|\G_{\tau-1}\right]\leq u$ for $\tau=1,2,\ldots,t$. Then, we have that
\begin{equation*} \label{new-1015-1}
\sum_{\tau=1}^tB_\tau\leq2\tc\gamma^\frac{1}{2}\ln(2/\delta)+\frac{3\gamma^\frac{1}{2}}{2\tc^\epsilon}\sum_{\tau=1}^{t}\beta_\tau^{1+\epsilon} u
\end{equation*}
holds with probability at least $1-T\delta$.
\end{lem}
\textbf{Proof.} First, we give the fact that
\begin{equation*}
\begin{aligned}
\left|\x_\tau^\top(\hat{\bt}_\tau-\bt_*)y_\tau\I_{|\beta_\tau y_\tau|\leq \tc}\right|\leq~&\norm{\hat{\bt}_\tau-\bt_*}_{\V_\tau}\norm{\x_\tau}_{\V_\tau^{-1}}|y_\tau|\I_{|\beta_\tau y_\tau|\leq \tc}\\
\leq~&\norm{\hat{\bt}_\tau-\bt_*}_{\V_\tau}\tc.
\end{aligned}
\end{equation*}
Then, through the full probability formula \citep{Mendenhall:2012}, we have that
\begin{equation}\label{ineq:0101-22}
\begin{aligned}
\P\left\{\sum_{\tau=1}^{t}B_\tau>\chi\right\}\leq~&\P\left\{\exists \tau, \norm{\hat{\bt}_\tau-\bt_*}^2_{\V_\tau}\geq\gamma\right\}+\P\left\{\sum_{\tau=1}^t B_\tau\I_{\norm{\hat{\bt}_\tau-\bt_*}^2_{\V_\tau}\leq\gamma}>\chi\right\}\\
 \leq~~& (T-1)\delta+\P\left\{\sum_{\tau=1}^t B_\tau\I_{\norm{\hat{\bt}_\tau-\bt_*}^2_{\V_\tau}\leq\gamma}>\chi\right\}.
\end{aligned}
\end{equation}
The second inequality of above equation holds because $\norm{\hat{\bt}_\tau-\bt_*}^2_{\V_\tau}\geq\gamma$ with probability at most $\delta$.

In the following, we analyze the term $\sum_{\tau=1}^t B_\tau\I_{\norm{\hat{\bt}_\tau-\bt_*}^2_{\V_\tau}\leq\gamma}$ to determine the appropriate $\chi$ for bounding the right side of \eqref{ineq:0101-22}. A simple application of the triangle inequality shows that 
\begin{equation}\label{0111-21}
\begin{aligned}
\sum_{\tau=1}^t B_\tau\I_{\norm{\hat{\bt}_\tau-\bt_*}^2_{\V_\tau}\leq\gamma}\leq~&\left|\sum_{\tau=1}^t \left(\tilde{y}_\tau-\E[\tilde{y}_\tau|\G_{\tau-1}]\right)\x_\tau^\top(\hat{\bt}_\tau-\bt_*)\I_{\norm{\hat{\bt}_\tau-\bt_*}^2_{\V_\tau}\leq\gamma}\right|\\
 +&\left|\sum_{\tau=1}^t \E[y_\tau\I_{|\beta_\tau y_\tau|\geq \tc}|\G_{\tau-1}]\x_\tau^\top(\hat{\bt}_\tau-\bt_*)\I_{\norm{\hat{\bt}_\tau-\bt_*}^2_{\V_\tau}\leq\gamma}\right|
\end{aligned}
\end{equation}
By utilizing Bernstein's inequality \citep[Lemma~11]{Bernstein}, we can demonstrate that,
\begin{equation*}
\begin{aligned}
&\left|\sum_{\tau=1}^t \left(\tilde{y}_\tau-\E[\tilde{y}_\tau|\G_{\tau-1}]\right)\x_\tau^\top(\hat{\bt}_\tau-\bt_*)\I_{\norm{\hat{\bt}_\tau-\bt_*}^2_{\V_\tau}\leq\gamma}\right|\\
\leq~&\frac{1}{2\tc\gamma^\frac{1}{2}}\sum_{\tau=1}^{t}\Var[\x_\tau^\top(\hat{\bt}_\tau-\bt_*)\I_{\norm{\hat{\bt}_\tau-\bt_*}^2_{\V_\tau}\leq\gamma}\tilde{y}_\tau|\G_{\tau-1}]+2\tc\gamma^\frac{1}{2}\ln(2/\delta).
\end{aligned}
\end{equation*}

holds with probability at least $1-\delta$. Additionally, apply the Cauchy-Schwarz inequality and the scalar property of variance, we can establish that
\begin{equation*}
\begin{aligned}
\Var[\x_\tau^\top(\hat{\bt}_\tau-\bt_*)\I_{\norm{\hat{\bt}_\tau-\bt_*}^2_{\V_\tau}\leq\gamma}\tilde{y}_\tau|\G_{\tau-1}]
\leq\gamma\cdot\Var[\beta_\tau \tilde{y}_\tau|\G_{\tau-1}].
\end{aligned}
\end{equation*}
Thus, we have that
\begin{equation}\label{0111-22}
\begin{aligned}
&\left|\sum_{\tau=1}^t \left(\tilde{y}_\tau-\E[\tilde{y}_\tau|\G_{\tau-1}]\right)\x_\tau^\top(\hat{\bt}_\tau-\bt_*)\I_{\norm{\hat{\bt}_\tau-\bt_*}^2_{\V_\tau}\leq\gamma}\right|\\
\leq~&\frac{\gamma}{2\tc\gamma^\frac{1}{2}}\sum_{\tau=1}^{t}\Var[\beta_\tau y_\tau\I_{|\beta_\tau y_\tau|\leq \tc}|\G_{\tau-1}]+2\tc\gamma^\frac{1}{2}\ln(2/\delta)\\
\leq~&\frac{\gamma^\frac{1}{2}}{2\tc^\epsilon}\sum_{\tau=1}^{t}\beta_\tau^{1+\epsilon} u+2\tc\gamma^\frac{1}{2}\ln(2/\delta).
\end{aligned}
\end{equation}
The second term on the right side of inequality \eqref{0111-21} can be bounded as 
\begin{equation}\label{0111-23}
\begin{aligned}
&\left|\sum_{\tau=1}^t \E[y_\tau\I_{|\beta_\tau y_\tau|\geq \tc}|\G_{\tau-1}]\x_\tau^\top(\hat{\bt}_\tau-\bt_*)\I_{\norm{\hat{\bt}_\tau-\bt_*}^2_{\V_\tau}\leq\gamma}\right|\\
\leq&\gamma^\frac{1}{2}\sum_{\tau=1}^t \E[|\beta_\tau y_\tau|\I_{|\beta_\tau y_\tau|\geq \tc}|\G_{\tau-1}]\leq\frac{\gamma^\frac{1}{2}}{\tc^\epsilon}\sum_{\tau=1}^{t}\beta_\tau^{1+\epsilon} u.
\end{aligned}
\end{equation}
Taking \eqref{0111-22}, \eqref{0111-23} into \eqref{0111-21}, we have the inequality
\begin{equation*}
\sum_{\tau=1}^t B_\tau\I_{\norm{\hat{\bt}_\tau-\bt_*}^2_{\V_\tau}\leq\gamma}\leq \frac{3\gamma^\frac{1}{2}}{2\tc^\epsilon}\sum_{\tau=1}^{t}\beta_\tau^{1+\epsilon} u+2\tc\gamma^\frac{1}{2}\ln(2/\delta).
\end{equation*}
holds with probability at least $1-\delta$. Let $\chi$ of inequality \eqref{ineq:0101-22} be $2\tc\gamma^\frac{1}{2}\ln(2/\delta)+\frac{3\gamma^\frac{1}{2}}{2\tc^\epsilon}\sum_{\tau=1}^{t}\beta_\tau^{1+\epsilon} u$, we have
\begin{equation*} \label{new-1015-2}
\P\left\{\sum_{\tau=1}^{t}B_\tau>\chi\right\}\leq T\delta.
\end{equation*}
The proof of Lemma~\ref{0519-2} is finished.
$\hfill\square$

We have bounded the terms $A$ and $\sum_{\tau=1}^{t}B_\tau$ using Lemma ~\ref{0519-1} and Lemma~\ref{0519-2}, respectively. By incorporating these two lemmas into equation \eqref{ineq:0101-19} and substituting $\delta$ with $\delta/2T$, we can derive that
\begin{equation}\label{new-1015-4}
\begin{aligned}
\norm{\hat{\bt}_{t+1}-\bt_*}_{\V_{t+1}}^2\leq~&6U^2\sum_{\tau=1}^t \beta_\tau^2+4\tc^2\ln(4T/\delta)+4\tc\gamma^\frac{1}{2}\ln(4T/\delta)\\
 &+ \lambda S^2+3\tc^{1-\epsilon}\sum_{\tau=1}^t \beta_\tau^{1+\epsilon}v+3\gamma^\frac{1}{2}\tc^{-\epsilon}\sum_{\tau=1}^{t}\beta_\tau^{1+\epsilon} u.
\end{aligned}
\end{equation}
holds with probability at least $1-\delta$. The H$\ddot{o}$lder  inequality tells that 
\begin{equation}\label{new-1015-3}
\sum_{\tau=1}^t \beta_\tau^{1+\epsilon}\leq t^\frac{1-\epsilon}{2}(\sum_{\tau=1}^t \beta_\tau^2)^\frac{1+\epsilon}{2}.
\end{equation}
Then, according to Lemma~11 of \cite{Abbasi:2011}, we have that
\begin{equation*}\label{ineq:27}
\begin{aligned}
\sum_{\tau=1}^T \beta_\tau^2=\sum_{\tau=1}^T\norm{\x_\tau}_{\V_\tau^{-1}}^2\leq\frac{4}{\kappa}\ln\left(\frac{\det(\V_{T+1})}{\det(\V_1)}\right)\leq\frac{4d}{\kappa}\ln\left(1+\frac{\kappa T}{2\lambda d}\right).
\end{aligned}
\end{equation*}
Thus, \eqref{new-1015-3} can be relaxed as

\begin{equation}\label{ineq:0519-1}
\begin{aligned}
\sum_{\tau=1}^t \beta_\tau^{1+\epsilon}\leq T^\frac{1-\epsilon}{2}\left(\frac{4d}{\kappa}\ln\left(1+\frac{\kappa T}{2\lambda d}\right)\right)^\frac{1+\epsilon}{2}.
\end{aligned}
\end{equation}

By taking \eqref{ineq:0519-1} into  \eqref{new-1015-4} and let 
\begin{equation*}
\tc=2(u\ln(4T/\delta))^\frac{1}{1+\epsilon}\left(d\kappa\ln\left(1+\frac{\kappa T}{2\lambda d}\right)\right)^\frac{1}{2}T^\frac{1-\epsilon}{2(1+\epsilon)},
\end{equation*} 
we have that
\begin{equation}\label{ineq:0519-7}
\begin{aligned}
\norm{\hat{\bt}_{t+1}-\bt_*}_{\V_{t+1}}^2\leq~& \lambda S^2+\frac{24U^2d}{\kappa}\ln\left(1+\frac{\kappa T}{2\lambda d}\right)\\
&+7u^\frac{2}{1+\epsilon}\ln(4T/\delta)^\frac{\epsilon-1}{1+\epsilon}T^\frac{1-\epsilon}{1+\epsilon}\frac{4d}{\kappa}\ln\left(1+\frac{\kappa T}{2\lambda d}\right)\\
 &+7u^\frac{1}{1+\epsilon}\ln(4T/\delta)^\frac{\epsilon}{1+\epsilon}T^\frac{1-\epsilon}{2(1+\epsilon)}\gamma^\frac{1}{2}\left(\frac{4d}{\kappa}\ln\left(1+\frac{\kappa T}{2\lambda d}\right)\right)^\frac{1}{2}
\end{aligned}
\end{equation}
holds with probability at least $1-\delta$. 

In order to determine $\gamma$ satisfying $\norm{\hat{\bt}_{t+1}-\bt_*}_{\V_{t+1}}^2\leq\gamma$, a quadratic inequality with respect to $\gamma$ need to be solved, such that the right side of inequality \eqref{ineq:0519-7} is smaller than $\gamma$. This leads to the conclusion that
\begin{equation*}
\begin{aligned}
\gamma=112v^\frac{2}{1+\epsilon}\ln(4T/\delta)^\frac{2\epsilon}{1+\epsilon}T^\frac{1-\epsilon}{1+\epsilon}\frac{4d}{\kappa}\ln\left(1+\frac{\kappa T}{2\lambda d}\right)+2\lambda S^2+\frac{48U^2d}{\kappa}\ln\left(1+\frac{\kappa T}{2\lambda d}\right).
\end{aligned}
\end{equation*}
By taking the union bound over all $t$, we have that with probability at least $1-\delta$, for any $t>0$, the inequality
\begin{equation*}
\begin{aligned}
\norm{\hat{\bt}_{t+1}-\bt_*}_{\V_{t+1}}^2\leq~&224v^\frac{2}{1+\epsilon}\ln(4T/\delta)^\frac{2\epsilon}{1+\epsilon}T^\frac{1-\epsilon}{1+\epsilon}\frac{4d}{\kappa}\ln\left(1+\frac{\kappa T}{2\lambda d}\right)\\
 &+2\lambda S^2+\frac{48U^2d}{\kappa}\ln\left(1+\frac{\kappa T}{2\lambda d}\right)
\end{aligned}
\end{equation*}
holds, which concludes the proof of Theorem~\ref{thm:3}.
$\hfill\square$

\section{Proof of Theorem \ref{thm:4}}

To begin with, we bound the instantaneous regret by the following lemma.
\begin{lem}\label{prop:1}
If $\bt_*\in\C_t$ for all $t$, then
\begin{equation*}
\mu(\tilde{\x}_t^\top\bt_*)-\mu(\x_t^\top\bt_*)\leq2L\sqrt{\gamma_t}\norm{\x_t}_{\V_t^{-1}}
\end{equation*}
where $\tilde{\x}_t=\argmax_{\x \in \D_t}\mu(\x^\top\bt_*)$.
\end{lem}

\textbf{Proof.} Considering that the link function $\mu(\cdot)$ is $L$-Lipschitz and monotonically increasing, we have
\begin{align*}
\mu(\tilde{\x}_t^\top\bt_*)-\mu(\x_t^\top\bt_*)\leq~&\max\{0, L(\tilde{\x}_t^\top\bt_*-\x_t^\top\bt_*)\}\\
\leq~&\max\{0, L(\x_t^\top\tilde{\bt}_t-\x_t^\top\bt_*)\}\\
=~&\max\{0, L\x_t^\top(\tilde{\bt}_t-\hat{\bt}_t)+L\x_t^\top(\hat{\bt}_t-\bt_*)\}\\
\leq~&L\left(\norm{\tilde{\bt}_t-\hat{\bt}_t}_{\V_t}+\norm{\hat{\bt}_t-\bt_*}_{\V_t}\right)\norm{\x_t}_{\V_t^{-1}}\\
\leq~& 2L\sqrt{\gamma_t}\norm{\x_t}_{\V_t^{-1}}
\end{align*}
where the second inequality holds due to the fact that $(\x_{t},\tilde{\bt}_{t})=\argmax_{\x\in\D_{t},\bt\in\C_{t}}\langle\x,\bt\rangle$.
$\hfill\square$

Then, we get the regret of CRTM through the cumulative summation from $1$ to $T$.

\begin{lem}\label{cor:1}
If $\bt_*\in\C_t$ for all $t$, then the regret of CRTM can be bounded as 
\begin{equation*}
R(T)\leq 2L\left(\frac{4d}{\kappa}\ln\left(1+\frac{\kappa T}{2\lambda d}\right)\sum_{t=1}^T\gamma_t\right)^{1/2}.
\end{equation*}
\end{lem}
\textbf{Proof.} Through the Lemma~\ref{prop:1}, we have that
\begin{equation}\label{regret:1}
\begin{aligned}
R(T)=~&\sum_{t=1}^T\mu(\tilde{\x}_t^\top\bt_*)-\mu(\x_t^\top\bt_*)\leq~2L\sum_{t=1}^T\sqrt{\gamma_t}\norm{\x_t}_{\V_t^{-1}}\\
\leq~&2L\left(\sum_{t=1}^T\gamma_t\right)^{1/2}\left(\sum_{t=1}^T\norm{\x_t}_{\V_t^{-1}}^2\right)^{1/2}.
\end{aligned}
\end{equation}
According to the Lemma~11 of \cite{Abbasi:2011}, we get that 
\begin{equation}\label{ineq:0519-8}
\begin{aligned}
\sum_{t=1}^T\norm{\x_t}_{\V_t^{-1}}^2\leq\frac{4}{\kappa}\ln\left(\frac{\det(\V_{T+1})}{\det(\V_1)}\right)\leq\frac{4d}{\kappa}\ln\left(1+\frac{\kappa T}{2\lambda d}\right).
\end{aligned}
\end{equation}
Combining \eqref{regret:1} and \eqref{ineq:0519-8} finishes the proof.
$\hfill\square$

By substituting $\gamma$ of Theorem~\ref{thm:3} into Lemma~\ref{cor:1} such that $\gamma_t=\gamma$ for $t=1, 2, \ldots, T$, the regret bound of CRTM is explicitly given as  
\begin{equation*}
\begin{aligned}
R(T)\leq~&128L\kappa^{-1}v^\frac{1}{1+\epsilon}d\ln(4T/\delta)^\frac{\epsilon}{1+\epsilon}\ln\left(1+\frac{\kappa T}{2\lambda d}\right)T^\frac{1}{1+\epsilon}\\
&+24LU\kappa^{-1}d\ln\left(1+\frac{\kappa T}{2\lambda d}\right)T^\frac{1}{2}\\
&+8LS(\lambda d)^\frac{1}{2}\kappa^{-\frac{1}{2}}\left(\ln\left(1+\frac{\kappa T}{2\lambda d}\right)\right)^\frac{1}{2}T^\frac{1}{2}\\
=~&O\left(d(\log T)^\frac{1+2\epsilon}{1+\epsilon}T^{\frac{1}{1+\epsilon}}\right).
\end{aligned}
\end{equation*}
The proof of Theorem \ref{thm:4} is finished.

\section{Proof of Theorem \ref{thm:1}}

Notice that CRMM updates the estimator with action-reward pair $(\x_t, \bar{y}_t)$, where $\bar{y}_t$ is the median of $\{y_t^1,y_t^2,\ldots,y_t^r\}$. Replace $(\x_t, y_t)$ of general upper bound \eqref{ineq:25} by $(\x_t, \bar{y}_t)$, we get that

\begin{equation*}
\begin{aligned}
\norm{\hat{\bt}_{t+1}-\bt_*}_{\V_{t+1}}^2\leq~& \norm{\hat{\bt}_1-\bt_*}_{\V_1}^2-\frac{\kappa}{2}\sum_{\tau=1}^t\alpha_\tau^2+\sum_{\tau=1}^t\left(\mu(\x_\tau^\top\bt_*)-\mu(\x_\tau^\top\hat{\bt}_\tau)\right)^2\norm{\x_\tau}_{\V_\tau^{-1}}^2\\
 &+\sum_{\tau=1}^t 2\x_\tau^\top(\hat{\bt}_\tau-\bt_*) \left(\bar{y}_\tau-\mu(\x_\tau^\top\bt_*)\right)+\sum_{\tau=1}^t \norm{\x_\tau}_{\V_\tau^{-1}}^2 \left(\bar{y}_\tau-\mu(\x_\tau^\top\bt_*)\right)^2.
\end{aligned}
\end{equation*}
Let $\alpha_\tau=\x_\tau^\top\left(\hat{\bt}_\tau-\bt_*\right), \beta_\tau=\norm{\x_\tau}_{\V_\tau^{-1}}$ and $X_\tau=\bar{y}_\tau-\mu(\x_\tau^\top\bt_*)$. The above equation can be simplified as

\begin{equation}\label{ineq-0519:25}
\begin{aligned}
\norm{\hat{\bt}_{t+1}-\bt_*}_{\V_{t+1}}^2\leq~& \norm{\hat{\bt}_1-\bt_*}_{\V_1}^2-\frac{\kappa}{2}\sum_{\tau=1}^t\alpha_\tau^2+\sum_{\tau=1}^t\left(\mu(\x_\tau^\top\bt_*)-\mu(\x_\tau^\top\hat{\bt}_\tau)\right)^2\norm{\x_\tau}_{\V_\tau^{-1}}^2\\
 &+\sum_{\tau=1}^t 2\alpha_\tau X_\tau+\sum_{\tau=1}^t \beta_\tau^2 X_\tau^2.
\end{aligned}
\end{equation}

We need to bound the terms $\sum_{\tau=1}^t \alpha_\tau X_\tau$ and $\sum_{\tau=1}^t \beta_\tau^2 X_\tau^2$ to conduct a narrow confidence region. Considering that the latent idea of CRMM is mean of medians, we provide the following lemma to display the $(1+\epsilon)$-th moment for the median term.

\begin{lem}\label{lem:7}
Suppose $X^1,\ldots ,X^r$ are independently drawn from the distribution $\chi$, and $\E[X^i]=0$, $\E[|X^i|^{1+\epsilon}]\leq v$ for $i=1,2,\ldots,r$. If $\widehat{X}$ is the median of $\{X^i\}_{i=1}^r$ , then $\widehat{X}$ satisfies $\E[|\widehat{X}|^{1+\epsilon}]\leq r v$.
\end{lem}
\textbf{Proof.} Let the p.d.f and c.d.f of $\chi$ be denoted as $p(x)$ and $F(x)$, respectively. Then, the c.d.f of $\widehat{X}$ can be calculated as
\begin{equation*}
\P\{\widehat{X}\leq x\}=\sum_{k=\lceil r/2 \rceil}^r\tbinom{r}{k}F(x)^k(1-F(x))^{r-k}.
\end{equation*}
Taking the derivative of the above equation, the p.d.f of $\widehat{X}$ can be obtained as
\begin{equation*}
f(x)=r\tbinom{r-1}{\lceil r/2 \rceil-1}F(x)^{\lceil r/2 \rceil-1}(1-F(x))^{r-\lceil r/2 \rceil}p(x).
\end{equation*}
According to the fact $\tbinom{r-1}{\lceil r/2 \rceil-1}F(x)^{\lceil r/2 \rceil-1}(1-F(x))^{r-\lceil r/2 \rceil}\leq1$, we can easily get that 
\begin{equation*}
f(x)\leq r p(x).
\end{equation*}
Thus, the $(1+\epsilon)$-th moment of $\widehat{X}$ satisfies
\begin{align*}
\E[|\widehat{X}|^{1+\epsilon}]= \int{|x|^{1+\epsilon}}f(x){\rm d} \leq r\int{|x|^{1+\epsilon}}p(x){\rm d} \leq r v.
\end{align*}
The proof of Lemma~\ref{lem:7} is finished.
$\hfill\square$

Another tool used to bound $\sum_{\tau=1}^t \alpha_\tau X_\tau$ is displayed as follows, whose  proof is provided in Section~\ref{proof-lem8}.
\begin{lem}\label{lem:8}
Suppose that $X_1,\ldots ,X_n$ are random variables satisfying $\E[X_i|\F_{i-1}]=0$, and $\E[|X_i|^{1+\epsilon}|\F_{i-1}]\leq v_1$, where $\F_{i-1}\triangleq\{X_1,\ldots,X_{i-1}\}$ is a $\sigma$-filtration and $\F_0=\emptyset$. For the fixed parameters $\alpha_1,\alpha_2,\ldots,\alpha_n\in\R$ and $C>0$, with probability at least $1-\delta$,  we have that
\begin{equation*}
\left|\sum_{i=1}^{n}\alpha_iX_i\mathbb{I}_{|\alpha_iX_i|\leq C\norm{\bm{\alpha}}_{1+\epsilon}}\right|\leq \xi\norm{\bm{\alpha}}_{1+\epsilon}
\end{equation*}
where 
\begin{align*}
\bm{\alpha}=[\alpha_1,\alpha_2,\ldots,\alpha_n],\xi=2C\ln(2/\delta)+2C^{-\epsilon}v_1.
\end{align*}
\end{lem}

Equipped with Lemma~\ref{lem:7} and Lemma~\ref{lem:8},  we are ready to bound the term $\sum_{\tau=1}^t \alpha_\tau X_\tau$.
\begin{lem}\label{lem:4}
Let $r=\left \lceil 16\ln\frac{4T}{\delta} \right \rceil$, for any $t>0$, with probability at least $1-\delta/T$,
\begin{equation*}
\sum_{\tau=1}^t \alpha_\tau X_\tau\leq\rho \norm{\bm{\alpha}}_{1+\epsilon}
\end{equation*}
where
\begin{align*}
\bm{\alpha}=[\alpha_1,\alpha_2,\ldots,\alpha_t], C=(4v)^{\frac{1}{1+\epsilon}}, \rho=2C\ln(4T/\delta)+2C^{-\epsilon}rv.
\end{align*}
\end{lem}
\textbf{Proof.} Through the full probability formula \citep{Mendenhall:2012}, we have that
\begin{equation}\label{eq:23}
\begin{aligned}
\Pr\left\{\left|\sum_{i=1}^{t}\alpha_\tau X_\tau\right|>\rho \norm{\bm{\alpha}}_{1+\epsilon}\right\}\leq~&\Pr\left\{\left|\sum_{\tau=1}^{t}\alpha_\tau X_\tau\mathbb{I}_{|\alpha_\tau X_\tau|\leq C \norm{\bm{\alpha}}_{1+\epsilon}}\right|>\rho \norm{\bm{\alpha}}_{1+\epsilon}\right\}\\
&+\sum_{\tau=1}^{t}\Pr\left\{|\alpha_\tau X_\tau|>C \norm{\bm{\alpha}}_{1+\epsilon}\right\}
\end{aligned}
\end{equation}
We first analyze the second term in the right side of \eqref{eq:23}. Recall that CRMM observes $r$ rewards $\{y_\tau^1,\ldots,y_\tau^r\}$ at round $\tau$, and for the sake of representation, we denote the difference between $y_\tau^i$ and $\mu(\x_\tau^\top\bt_*)$ as $X_\tau^i$, such that $X_\tau^i=y_\tau^i-\mu(\x_\tau^\top\bt_*)$. Through Markov's inequality and the heavy-tailed condition $\E[|X_\tau^i|^{1+\epsilon}]\leq v$, we have that
\begin{equation*}
\begin{aligned}
\Pr\left\{|\alpha_\tau X_\tau^i|>C \norm{\bm{\alpha}}_{1+\epsilon}\right\}\leq\frac{|\alpha_\tau|^{1+\epsilon}v}{C ^{1+\epsilon}\norm{\bm{\alpha}}_{1+\epsilon}^{1+\epsilon}}.
\end{aligned}
\end{equation*}
Let $C=(4v)^{\frac{1}{1+\epsilon}}$, then we have 
\begin{equation*}
\Pr\left\{|\alpha_\tau X_\tau^i|>C \norm{\bm{\alpha}}_{1+\epsilon}\right\}\leq\frac{1}{4}.
\end{equation*}
Define the random variables
\begin{equation*}
B_i=\mathbbm{I}_{\alpha_\tau X_\tau^i>C \norm{\bm{\alpha}}_{1+\epsilon}},
\end{equation*}
thus $p_i=\Pr\{B_i=1\}\leq\frac{1}{4}$. According to the Azuma-Hoeffing's inequality \citep{azuma1967}, we get
\begin{align*}
\Pr\left\{\sum_{i=1}^{r}B_j\geq\frac{r}{2}\right\}\leq&\Pr\left\{\sum_{i=1}^{r}B_i-p_i\geq \frac{r}{4}\right\}\\
\leq &e^{-r/8}\leq\frac{\delta}{4T^2}
\end{align*}
for $r=\left \lceil 16\ln\frac{4T}{\delta} \right \rceil$. The inequality $\sum_{i=1}^{r}B_i\geq \frac{r}{2}$ means more than half of the terms $\{B_i\}_{i=1}^{r}$ is true. Thus, the median term $\alpha_\tau X_\tau$ satisfies 
\begin{equation*}
\alpha_\tau X_\tau>C \norm{\bm{\alpha}}_{1+\epsilon}
\end{equation*} 
with probability at most $\frac{\delta}{4T^2}$. A similar argument shows that 
\begin{equation*}
\alpha_\tau X_\tau<-C \norm{\bm{\alpha}}_{1+\epsilon}
\end{equation*}
holds with probability at most $\frac{\delta}{4T^2}$. Therefore, we have
\begin{equation*}
\Pr\left\{|\alpha_\tau X_\tau|>C \norm{\bm{\alpha}}_{1+\epsilon}\right\}\leq \frac{\delta}{2T^2}.
\end{equation*}
By taking it into \eqref{eq:23}, we have that
\begin{equation}\label{eq:24}
\begin{aligned}
\Pr\left\{\left|\sum_{\tau=1}^{t}\alpha_\tau X_\tau\right|>\rho \norm{\bm{\alpha}}_{1+\epsilon}\right\}\leq\frac{\delta}{2T}+\Pr\left\{\left|\sum_{\tau=1}^{t}\alpha_\tau X_\tau\mathbb{I}_{|\alpha_\tau X_\tau|\leq C \norm{\bm{\alpha}}_{1+\epsilon}}\right|>\rho \norm{\bm{\alpha}}_{1+\epsilon}\right\}.
\end{aligned}
\end{equation}

Next, we proceed to bound the second term on the right side of inequality \eqref{eq:24} using Lemma~\ref{lem:8}. The application of Lemma~\ref{lem:8} requires satisfying two conditions. The first condition is that the expectation of the median term $X_\tau$ is $0$, which is easily fulfilled due to the symmetry of rewards. The second condition is that the $(1+\epsilon)$-th moment of $X_\tau$ is finite, which can be verified through Lemma~\ref{lem:7}, such that
\begin{equation*}
\E[|X_\tau|^{1+\epsilon}]\leq r v.
\end{equation*}
Consequently, Lemma~\ref{lem:8} can be employed to bound the second term on the right side of \eqref{eq:24} by setting $C=(4v)^{\frac{1}{1+\epsilon}}$ and $v_1=rv$. This yields that
\begin{equation*}
\left|\sum_{i=1}^t\alpha_\tau X_\tau\right|\leq (2C\ln(4T/\delta)+2C^{-\epsilon}rv) \norm{\bm{\alpha}}_{1+\epsilon}
\end{equation*}
holds with probability at least $1-\delta/T$. Hence, the proof of Lemma~\ref{lem:4} is concluded.
$\hfill\square$

Similar to the discussion of Lemma~\ref{lem:4}, we present Lemma~\ref{lem:5} to bound the term $\sum_{\tau=1}^t \beta_\tau^2 X_\tau^2$. The proof of Lemma~\ref{lem:5} is provided in Section~\ref{proof-lem5}.

\begin{lem}\label{lem:5}
Let $r=\left \lceil 16\ln\frac{4T}{\delta} \right \rceil$, for any $t>0$, with probability at least $1-\delta/T$, 
\begin{equation*}
\sum_{\tau=1}^t\beta_\tau^2X_\tau^2\leq C\rho \norm{\bm{\beta}}_{1+\epsilon}^2
\end{equation*}
where
\begin{align*}
\bm{\beta}=[\beta_1,\beta_2,\ldots,\beta_t], C=(4v)^{\frac{1}{1+\epsilon}}, \rho=2C\ln(4T/\delta)+2C^{-\epsilon}rv.
\end{align*}
\end{lem}
By taking Lemma~\ref{lem:4} and Lemma~\ref{lem:5} into inequality \eqref{ineq-0519:25}, we get that
\begin{equation}\label{ineq:26}
\begin{aligned}
\norm{\hat{\bt}_{t+1}-\bt_*}_{\V_{t+1}}^2\leq~&\norm{\hat{\bt}_1-\bt_*}_{\V_1}^2+\sum_{\tau=1}^t\left(\mu(\x_\tau^\top\bt_*)-\mu(\x_\tau^\top\hat{\bt}_\tau)\right)^2\norm{\x_\tau}_{\V_\tau^{-1}}^2\\
 &-\frac{\kappa}{2}\norm{\bm{\alpha}}_2^2+2\rho \norm{\bm{\alpha}}_{1+\epsilon}+C\rho \norm{\bm{\beta}}_{1+\epsilon}^2
\end{aligned}
\end{equation}
holds with probability at least $1-2\delta/T$.

Recall the upper bound of $\mu(\cdot)$ is $U$ and $\norm{\hat{\bt}_1-\bt_*}_{\V_1}^2\leq \lambda S^2$, inequality \eqref{ineq:26} can be simplified as 

\begin{equation}\label{ineq:262}
\begin{aligned}
\norm{\hat{\bt}_{t+1}-\bt_*}_{\V_{t+1}}^2\leq\lambda S^2+4U^2\sum_{\tau=1}^t\norm{\x_\tau}_{\V_\tau^{-1}}^2-\frac{\kappa}{2}\norm{\bm{\alpha}}_2^2+2\rho \norm{\bm{\alpha}}_{1+\epsilon}+C\rho \norm{\bm{\beta}}_{1+\epsilon}^2
\end{aligned}
\end{equation}
Based on the H$\ddot{o}$lder inequality, we get that
\begin{align*}
\norm{\bm{\alpha}}_{1+\epsilon}\leq t^{\frac{1-\epsilon}{2(1+\epsilon)}}\norm{\bm{\alpha}}_2,\ 
\norm{\bm{\beta}}_{1+\epsilon}^2\leq t^{\frac{1-\epsilon}{1+\epsilon}}\norm{\bm{\beta}}_2^2.
\end{align*}
By taking these two inequalities into \eqref{ineq:262} and recalling that $\beta_\tau=\norm{\x_\tau}_{\V_\tau^{-1}}$, we get that
\begin{align*}
\norm{\hat{\bt}_{t+1}-\bt_*}_{\V_{t+1}}^2\leq~&\lambda S^2+\left(4U^2+C\rho t^{\frac{1-\epsilon}{1+\epsilon}}\right)\sum_{\tau=1}^t\norm{\x_\tau}_{\V_\tau^{-1}}^2-\frac{\kappa}{2}\norm{\bm{\alpha}}_2^2+2\rho t^{\frac{1-\epsilon}{2(1+\epsilon)}}\norm{\bm{\alpha}}_2.
\end{align*}
holds with probability at least $1-2\delta/T$. 

According to the fact $2\sqrt{pq}\leq\frac{p}{\kappa}+\kappa q,\forall p,q>0$, if we take $p=4\rho^2 t^{\frac{1-\epsilon}{1+\epsilon}}, q=\norm{\bm{\alpha}}_2^2$, we get that
\begin{align*}
\norm{\hat{\bt}_{t+1}-\bt_*}_{\V_{t+1}}^2\leq\left(4U^2+C\rho t^{\frac{1-\epsilon}{1+\epsilon}}\right)\sum_{\tau=1}^t\norm{\x_\tau}_{\V_\tau^{-1}}^2+\lambda S^2+\frac{2\rho^2}{\kappa} t^{\frac{1-\epsilon}{1+\epsilon}}.
\end{align*}
holds with probability at least $1-2\delta/T$. Then, take an union bound over all rounds and , we have that with probability at least $1-2\delta$, for any $t>0$,
\begin{align*}
\norm{\hat{\bt}_{t+1}-\bt_*}_{\V_{t+1}}^2 \leq&\left(4U^2+C\rho t^{\frac{1-\epsilon}{1+\epsilon}}\right)\frac{4d}{\kappa}\ln\left(1+\frac{\kappa t}{2\lambda d}\right)+\lambda S^2+\frac{2\rho^2}{\kappa} t^{\frac{1-\epsilon}{1+\epsilon}}.
\end{align*}
The proof of Theorem \ref{thm:1} is finished.

\section{Proof of Theorem \ref{thm:2}}

Since CRMM plays total $T_0$ rounds with $T_0=\lfloor T/r \rfloor$ and $r=\left \lceil 16\ln\frac{4T}{\delta} \right \rceil$, we bound the sum of $\gamma_t$ from $t=1$ to $T_0$ first, such that
\begin{equation*}
\begin{aligned}
\sum_{t=1}^{T_0}\gamma_t\leq~& \left(\frac{16U^2d}{\kappa}\ln\left(1+\frac{\kappa T_0}{2\lambda d}\right)+\lambda S^2\right)T_0+\left(\frac{2\rho^2}{\kappa}T_0+\frac{4C\rho d}{\kappa}\ln\left(1+\frac{\kappa T_0}{2\lambda d}\right)\right)\sum_{t=1}^{T_0}t^{\frac{1-\epsilon}{1+\epsilon}}\\
\leq~&\left(\frac{16U^2d}{\kappa}\ln\left(1+\frac{\kappa T_0}{2\lambda d}\right)+\lambda S^2\right)T_0+\left(\frac{2\rho^2}{\kappa}+\frac{4C\rho d}{\kappa}\ln\left(1+\frac{\kappa T_0}{2\lambda d}\right)\right)T_0^{\frac{2}{1+\epsilon}}.
\end{aligned}
\end{equation*}
The second inequality holds due to the fact $\sum_{t=1}^{T_0}t^{\frac{1-\epsilon}{1+\epsilon}}\leq\int_{0}^{T_0} x^{\frac{1-\epsilon}{1+\epsilon}}\mathrm{d}x\leq {T_0}^{\frac{2}{1+\epsilon}}$. Taking above result into Lemma~\ref{cor:1}, we can easily get that
\begin{equation*}
\begin{aligned}
R(T_0)\leq&~16LUd\kappa^{-1}\ln\left(1+\frac{\kappa T_0}{2\lambda d}\right)T_0^\frac{1}{2}+4LS\kappa^{-\frac{1}{2}}\left(\lambda d\ln\left(1+\frac{\kappa T_0}{2\lambda d}\right)\right)^\frac{1}{2}T_0^\frac{1}{2}\\
&+8L\rho\kappa^{-1}\left(d\ln\left(1+\frac{\kappa T_0}{2\lambda d}\right)\right)^\frac{1}{2}T_0^{\frac{1}{1+\epsilon}}+8Ld\kappa^{-1}C\rho^\frac{1}{2}\ln\left(1+\frac{\kappa T_0}{2\lambda d}\right)T_0^{\frac{1}{1+\epsilon}}.
\end{aligned}
\end{equation*}
Taking $R(T)=rR(T_0)$ shows that the regret of CRMM can be bounded as
\begin{equation*}
\begin{aligned}
R(T)\leq~& 64LUd\kappa^{-1}\ln\left(1+\frac{\kappa T}{2\lambda d}\right)\left(\ln\frac{4T}{\delta}\right)^\frac{1}{2}T^\frac{1}{2}\\
&+16LS\kappa^{-\frac{1}{2}}\left(\lambda d\ln\left(1+\frac{\kappa T}{2\lambda d}\right)\ln\frac{4T}{\delta}\right)^\frac{1}{2}T^\frac{1}{2}\\
&+32L\rho\kappa^{-1}\left(d\ln\left(1+\frac{\kappa T}{2\lambda d}\right)\right)^\frac{1}{2}\left(\ln\frac{4T}{\delta}\right)^\frac{\epsilon}{1+\epsilon}T^{\frac{1}{1+\epsilon}}\\
&+32Ld\kappa^{-1}C\rho^\frac{1}{2}\ln\left(1+\frac{\kappa T}{2\lambda d}\right)\left(\ln\frac{4T}{\delta}\right)^\frac{\epsilon}{1+\epsilon}T^{\frac{1}{1+\epsilon}}\\
=~&O\left(d(\log T)^{\frac{3}{2}+\frac{\epsilon}{1+\epsilon}}T^{\frac{1}{1+\epsilon}}\right).
\end{aligned}
\end{equation*}
The proof of Theorem \ref{thm:2} is finished.

\section{Proof of Lemma \ref{lem:8}} \label{proof-lem8}
Let $Z_i=X_i\mathbb{I}_{|\alpha_iX_i|\leq C\norm{\bm{\alpha}}_{1+\epsilon}}$. Based on the triangle inequality, we obtain that
\begin{equation}\label{inequality:21}
\begin{aligned}
\left|\sum_{i=1}^{n}\alpha_iZ_i\right|\leq\left|\sum_{i=1}^{n}\alpha_iZ_i -\E\left[\alpha_iZ_i |\F_{i-1}\right]\right|+\left|\sum_{i=1}^{n}\E\left[\alpha_iZ_i |\F_{i-1}\right]\right|.
\end{aligned}
\end{equation}

Utilizing Bernstein's inequality \citep[Lemma~11]{Bernstein} for the first term in the right side of obove equation shows that with probability at least $1-\delta$, we have
\begin{equation*}
\left|\sum_{i=1}^{n}\alpha_iZ_i -\E\left[\alpha_iZ_i |\F_{i-1}\right]\right|\leq2C\norm{\bm{\alpha}}_{1+\epsilon}\ln(2/\delta)+\frac{1}{2C\norm{\bm{\alpha}}_{1+\epsilon}}\sum_{i=1}^n\Var[\alpha_iZ_i|\F_{i-1}].
\end{equation*}
The variance of $Z_i$ can be relaxed as follows,
\begin{align*}
\sum_{i=1}^n\Var[\alpha_iZ_i|\F_{i-1}] =~&\sum_{i=1}^{n}\E\left[(\alpha_iZ_i -\E[\alpha_iZ_i|\F_{i-1}])^2|\F_{i-1}\right]\\
\leq~& \sum_{i=1}^{n}\E\left[(\alpha_iZ_i)^2|\F_{i-1}\right]\leq vC^{1-\epsilon}\norm{\bm{\alpha}}_{1+\epsilon}^2.
\end{align*}
Thus, we get that
\begin{equation}\label{ineq:0519-2}
\left|\sum_{i=1}^{n}\alpha_iZ_i -\E\left[\alpha_iZ_i |\F_{i-1}\right]\right|\leq (2C\ln(2/\delta)+vC^{-\epsilon})\norm{\bm{\alpha}}_{1+\epsilon}
\end{equation}
holds with probability at least $1-\delta$.

According to the conditions $\E[X_i|\F_{i-1}]=0$ and $\E[|X_i|^{1+\epsilon}|\F_{i-1}]\leq v$ for $i=1,2,\ldots,n$, we can easily obtain that
\begin{equation}\label{ineq:0519-3}
\begin{aligned}
\left|\sum_{i=1}^{n}\E\left[\alpha_iZ_i |\F_{i-1}\right]\right|=~&\left|\sum_{i=1}^{n}\E\left[\alpha_iX_i\mathbb{I}_{|\alpha_iX_i|\leq C\norm{\bm{\alpha}}_{1+\epsilon}} |\F_{i-1}\right]\right|\\
\leq~&\sum_{i=1}^{n}\E\left[|\alpha_iX_i|\mathbb{I}_{|\alpha_iX_i|> C\norm{\bm{\alpha}}_{1+\epsilon}} |\F_{i-1}\right]\\
\leq~&\sum_{i=1}^{n} \left(\E\left[|\alpha_iX_i|^{1+\epsilon}|\F_{i-1}\right]\right)^\frac{1}{1+\epsilon}\Pr\left\{|\alpha_iX_i|>C\norm{\bm{\alpha}}_{1+\epsilon}\right\}^{\frac{\epsilon}{1+\epsilon}}\\
=~& vC^{-\epsilon}\norm{\bm{\alpha}}_{1+\epsilon}.
\end{aligned}
\end{equation}
Taking \eqref{ineq:0519-2} and \eqref{ineq:0519-3} into \eqref{inequality:21} finishes the proof.

\section{Proof of Lemma \ref{lem:5}} \label{proof-lem5}
We first provide the following lemma to help with the proof of Lemma \ref{lem:5}. 

\begin{lem}\label{lem:10}
Let $X_1,\ldots ,X_n$ be random variables with bounded moments $\E[|X_i|^{1+\epsilon}|\F_{i-1}]\leq v_1$, where $\F_{i-1}\triangleq\{X_1,\ldots,X_{i-1}\}$ is a $\sigma$-filtration and $\F_0=\emptyset$. For the fixed parameters $\beta_1,\beta_2,\ldots,\beta_n\in\R$ and $C>0$, with probability at least $1-\delta$,  we have that
\begin{equation*}
\sum_{i=1}^{n}\beta_i^2X_i^2\mathbb{I}_{\beta_i^2X_i^2\leq C^2\norm{\bm{\beta}}_{1+\epsilon}^2}\leq \xi\norm{\bm{\beta}}_{1+\epsilon}^2,
\end{equation*}
where 
\begin{align*}
\bm{\beta}=[\beta_1,\beta_2,\ldots,\beta_n],\xi=2C^2\ln(2/\delta)+2v_1C^{1-\epsilon}.
\end{align*}
\end{lem}

\textbf{Proof.} Let $Z_i^2=X_i^2\mathbb{I}_{\beta_i^2X_i^2\leq C^2\norm{\bm{\beta}}_{1+\epsilon}^2}$. The triangle inequality shows that
\begin{equation}\label{inequality:10}
\begin{aligned}
\sum_{i=1}^{n}\beta_i^2Z_i^2\leq\left|\sum_{i=1}^{n}\beta_i^2Z_i^2-\E\left[\beta_i^2Z_i^2 |\F_{i-1}\right]\right|+\left|\sum_{i=1}^{n}\E\left[\beta_i^2Z_i^2 |\F_{i-1}\right]\right|.
\end{aligned}
\end{equation}
Taking use of the Bernstein's inequality \citep[Lemma~11]{Bernstein} tells that
\begin{equation*}
 \left|\sum_{i=1}^{n}\beta_i^2Z_i^2-\E\left[\beta_i^2Z_i^2 |\F_{i-1}\right]\right|\leq2C^2\norm{\bm{\beta}}_{1+\epsilon}^2\ln(2/\delta)+\frac{1}{2C^2\norm{\bm{\beta}}_{1+\epsilon}^2}\sum_{i=1}^n\Var[\beta_i^2Z_i^2|\F_{i-1}]
\end{equation*}
holds with probability at least $1-\delta$. The variance of $\beta_i^2Z_i^2$ can be relaxed as 
\begin{equation*}
\sum_{i=1}^{n}\E\left[(\beta_i^2Z_i^2 -\E[\beta_i^2Z_i^2])^2|\F_{i-1}\right]\leq \sum_{i=1}^{n}\E\left[(\beta_iZ_i)^4|\F_{i-1}\right]\leq v_1C^{3-\epsilon}\norm{\bm{\beta}}_{1+\epsilon}^4.
\end{equation*}
Thus, we get that
\begin{equation}\label{inequality:011}
 \left|\sum_{i=1}^{n}\beta_i^2Z_i^2-\E\left[\beta_i^2Z_i^2 |\F_{i-1}\right]\right|\leq2C^2\norm{\bm{\beta}}_{1+\epsilon}^2\ln(2/\delta)+v_1C^{1-\epsilon}\norm{\bm{\beta}}_{1+\epsilon}^2.
\end{equation}

Considering that $\E[|X_i|^{1+\epsilon}|\F_{i-1}]\leq v_1$, $i=1,2,\ldots,n$, it is easy to verify that
\begin{equation}\label{ineq:0519-5}
\sum_{i=1}^{n}\E\left[\beta_i^2Z_i^2 |\F_{i-1}\right]\leq v_1C^{1-\epsilon}\norm{\bm{\beta}}_{1+\epsilon}^2.
\end{equation}
Taking \eqref{inequality:011} and \eqref{ineq:0519-5} into \eqref{inequality:10} finishes the proof of Lemma \ref{lem:10}.
$\hfill\square$

Now, we are ready to prove Lemma \ref{lem:5}. Through the full probability formula \citep{Mendenhall:2012}, we have that
\begin{equation}\label{eq:center-mom}
\begin{aligned}
 \Pr\left\{\sum_{\tau=1}^{t}\beta_\tau^2X_\tau^2>C\rho \norm{\bm{\beta}}_{1+\epsilon}^2\right\}\leq&\Pr\left\{\sum_{\tau=1}^{t}\beta_\tau^2X_\tau^2\mathbb{I}_{\beta_\tau^2X_\tau^2\leq C^2 \norm{\bm{\beta}}_{1+\epsilon}^2}>C\rho \norm{\bm{\beta}}_{1+\epsilon}^2\right\}\\
&+\sum_{\tau=1}^{t}\Pr\left\{|\beta_\tau X_\tau|>C \norm{\bm{\beta}}_{1+\epsilon}\right\}.
\end{aligned}
\end{equation}
We first analyze the second term on the right side of above inequality. Recall that CRMM observes $r$ rewards $\{y_t^1,\ldots,y_t^r\}$ at round $t$, and for the sake of representation, we denote the difference between $y_t^i$ and $\mu(\x_t^\top\bt_*)$ as $X_t^i$, such that $X_t^i=y_t^i-\mu(\x_t^\top\bt_*)$. Through Markov's inequality and the heavy-tailed condition $\E[|X_\tau|^{1+\epsilon}]\leq v$, we have that
\begin{equation*}
\begin{aligned}
\Pr\left\{|\beta_\tau X_\tau^i|>C \norm{\bm{\beta}}_{1+\epsilon}\right\}\leq\frac{|\beta_\tau|^{1+\epsilon}v}{C ^{1+\epsilon}\norm{\bm{\beta}}_{1+\epsilon}^{1+\epsilon}}.
\end{aligned}
\end{equation*}
Let $C=(4v)^{\frac{1}{1+\epsilon}}$, then we have 
\begin{equation*}
\Pr\left\{|\beta_\tau X_\tau^i|>C \norm{\bm{\beta}}_{1+\epsilon}\right\}\leq\frac{1}{4}.
\end{equation*}
Define the random variables
\begin{equation*}
B_i=\mathbbm{I}_{\beta_\tau X_\tau^i>C \norm{\bm{\beta}}_{1+\epsilon}},
\end{equation*}
thus $p_i=\Pr\{B_i=1\}\leq\frac{1}{4}$. According to the Azuma-Hoeffing's inequality \citep{azuma1967}, we have that
\begin{align*}
\Pr\left\{\sum_{i=1}^{r}B_j\geq\frac{r}{2}\right\}\leq&\Pr\left\{\sum_{i=1}^{r}B_i-p_i\geq \frac{r}{4}\right\}\\
\leq &e^{-r/8}\leq\frac{\delta}{4T^2}
\end{align*}
for $r=\left \lceil 16\ln\frac{4T}{\delta} \right \rceil$. The inequality $\sum_{i=1}^{r}B_i\geq \frac{r}{2}$ means more than half of the terms $\{B_i\}_{i=1}^{r}$ is true. Thus, the median term $\beta_\tau X_\tau$ satisfies 
\begin{equation*}
\beta_\tau X_\tau>C \norm{\bm{\beta}}_{1+\epsilon}
\end{equation*} 
with probability at most $\frac{\delta}{4T^2}$. A similar argument shows that 
\begin{equation*}
\beta_\tau X_\tau<-C \norm{\bm{\beta}}_{1+\epsilon}
\end{equation*}
holds with probability at most $\frac{\delta}{4T^2}$. Therefore, we have
\begin{equation*}
\Pr\left\{|\beta_\tau X_\tau|>C \norm{\bm{\beta}}_{1+\epsilon}\right\}\leq \frac{\delta}{2T^2}.
\end{equation*}
Take it into \eqref{eq:center-mom}, we get that
\begin{equation*}
\begin{aligned}
\Pr\left\{\sum_{\tau=1}^{t}\beta_\tau^2X_\tau^2>C\rho \norm{\bm{\beta}}_{1+\epsilon}\right\}\leq\frac{\delta}{2T}+\Pr\left\{\sum_{\tau=1}^{t}\beta_\tau^2X_\tau^2\mathbb{I}_{\beta_\tau^2X_\tau^2\leq C^2 \norm{\bm{\beta}}_{1+\epsilon}^2}>C\rho \norm{\bm{\beta}}_{1+\epsilon}^2\right\}.
\end{aligned}
\end{equation*}
We obtain that the $(1+\epsilon)$-th moment of $X_\tau$ is $rv$ by Lemma~\ref{lem:7}, thus Lemma~\ref{lem:10} can be taken to bound the second term on the right side of above inequality, such that 
\begin{equation*}
\Pr\left\{\sum_{\tau=1}^{t}\beta_\tau^2X_\tau^2\mathbb{I}_{\beta_\tau^2X_\tau^2\leq C^2 \norm{\bm{\beta}}_{1+\epsilon}^2}>C\rho \norm{\bm{\beta}}_{1+\epsilon}^2\right\}\leq \frac{\delta}{2T}
\end{equation*}
with $\rho=2C\ln(4T/\delta)+2C^{-\epsilon}rv$. Thus, we get that
\begin{equation*}
\sum_{\tau=1}^t\beta_\tau^2X_\tau^2\leq (2C^2\ln(4T/\delta)+2r vC^{1-\epsilon}) \norm{\bm{\beta}}_{1+\epsilon}^2
\end{equation*}
holds with probability at least $1-\delta/T$.

\end{document}